\documentclass[journal]{IEEEtran}
\ifCLASSINFOpdf
\else
\fi

\usepackage{amsmath,amsfonts,amssymb,latexsym,epic,epsfig,graphics,psfrag,booktabs}
\usepackage{subfigure}
\usepackage{algorithm}               %format of the algorithm
\usepackage{algorithmic}             %//format of the algorithm
\usepackage{cite}
\usepackage{multirow}
\usepackage{graphicx}
\usepackage{epsfig}
\usepackage{epstopdf}

\begin{document}
\title{A Hierarchical Image Matting Model for Blood Vessel Segmentation in Fundus images}

\author{Zhun Fan, \IEEEmembership{Senior Member,~IEEE,}
        Jiewei Lu,
        Wenji Li, \\
        Caimin Wei,
        Han Huang,
        Xinye Cai,
        Xinjian Chen$^*$, \IEEEmembership{Senior Member,~IEEE,}
\thanks{

This work was supported by the National Natural Foundation of Guangdong Province, Integrated Platform
 of Evolutionary Intelligence and Robot, support no.(2015KGJHZ015).

Zhun Fan, Jiewei Lu and Wenji Li are with the Guangdong
Provincial Key Laboratory of Digital Signal and Image Processing, College of
Engineering, Shantou University, Shan'tou 515063, China (email: zfan, 12jwlu1, liwj@stu.edu.cn).

Caimin Wei is with the Department of Mathematics, Shantou University, Shan'tou 515063, China (email: cmwei@stu.edu.cn).

Han Huang is with the School of Software Engineering, South China University of Technology, Guang'zhou 510006, China (email: hhan@scut.edu.cn).

Xinye Cai is with the College of Computer Science and Technology, Nanjing University of Aeronautics and Astronautics, Jiang'su 210016, China (email:xinye@nuaa.edu.cn).

Xinjian Chen is with the Medical Image Processing and Analysis Lab, School of Electronics and Information Engineering, Soochow University, Su'zhou 215006, China (email: xjchen@suda.edu.cn).

}% <-this % stops a space

}


\markboth{}%
{Shell \MakeLowercase{\textit{et al.}}: Bare Demo of IEEEtran.cls for IEEE Journals}



\maketitle

\begin{abstract}
In this paper, a hierarchical image matting model is proposed to extract blood vessels from fundus images. More specifically, a hierarchical strategy utilizing the continuity and extendibility of retinal blood vessels is integrated into the image matting model for blood vessel segmentation. Normally the matting models require the user specified \emph{trimap}, which separates the input image into three regions manually: the foreground, background and unknown regions. However, since creating a user specified trimap is a tedious and time-consuming task, region features of blood vessels are used to generate the trimap automatically in this paper. The proposed model has low computational complexity and outperforms many other state-of-art supervised and unsupervised methods in terms of accuracy, which achieves a vessel segmentation accuracy of $96.0\%$, $95.7\%$ and $95.1\%$ in an average time of $10.72s$, $15.74s$ and $50.71s$ on images from three publicly available fundus image datasets DRIVE, STARE, and CHASE\_DB1, respectively.
\end{abstract}


% \fi
\IEEEpeerreviewmaketitle

\section{Introduction}
\IEEEPARstart{R}{etinal} blood vessels generally show a coarse to fine centrifugal distribution and appear as a wire mesh-like structure or tree-like structure. Their morphological features, such as length, width and branching, play an important role in diagnosis, screening, early detection and treatment of various cardiovascular and ophthalmologic diseases such as stroke, vein occlusions, diabetes and arteriosclerosis \cite{kanski2011clinical}. The analysis of morphological features of retinal blood vessels can facilitate a timely detection and treatment of a disease when it is still in its early stage. Moreover, the analysis of retinal blood vessels can assist in evaluation of retinal image registration \cite{zana1999multimodal}, the relationship between vessel tortuosity and hypertensive retinopathy \cite{foracchia2001extraction},  retinopathy of prematurity \cite{heneghan2002characterization},  arteriolar narrowing \cite{grisan2003divide}, mosaic synthesis \cite{fritzsche2003automated}, biometric identification \cite{marino2006personal}, foveal avascular region detection \cite{haddouche2010detection} and computer-assisted laser surgery \cite{kanski2011clinical}. Since cardiovascular and ophthalmologic diseases have a serious impact on human's life, the analysis of retinal blood vessels becomes more and more important. It is of great significance in many clinical applications to reveal important information of systemic diseases and support diagnosis and treatment. As a result, the requirement of vessel analysis system grows rapidly in which the segmentation of retinal blood vessels is the first and one of the most crucial steps.

The segmentation of retinal blood vessels has been a heavily researched area in recent years \cite{fraz2012blood}. Broadly speaking, existing algorithms can be divided into supervised and unsupervised methods. In supervised methods, a number of different features are extracted from fundus images, and applied to train the effective classifiers with the purpose of extracting retinal blood vessels. In \cite{staal2004ridge}, Staal \emph{et al.} extracts 27 features for each image pixel with ridge profiles, and performs feature selection by using sequential forward selection method to pick those pixels that result in better segmentation performance by a K-Nearest Neighbor (KNN) classifier. Soares \emph{et al.} \cite{soares2006retinal} introduces a feature-based Bayesian classifier with Gaussian mixtures for vessel segmentation, which uses the intensity information and Gabor wavelet transform responses to build a 7-D feature vector for each pixel. In \cite{lupascu2010fabc}, Lupascu \emph{et al.} utilizes an AdaBoost classifier and a 41-D feature vector which includes information on the local intensity structure, spatial properties, and geometry at multiple scales. Marin \emph{et al.} \cite{marin2011new} constructs a 7-D vector composed of gray-level and moment invariants-based features, and then trains a neural network (NN) for pixel classification. Roychowdhury \emph{et al.}\cite{roychowdhury2015blood} extracts the major vessel from the fundus images and uses a Gaussian Mixture Model classifier for vessel segmentation with a set of 8 features, which are extracted based on pixel neighborhood and first and second-order gradient images. In \cite{Liskowski2016Segmenting}, Liskowski \emph{et al.} employs a deep neural network to extract vessel pixels from fundus images. In unsupervised methods, the researchers try to find inherent properties of retinal blood vessels that can be applied to extract vessel pixels from fundus images. The unsupervised methods can be further divided into multiscale approaches, matched filtering, vessel tracking, mathematical morphology and model based methods \cite{fraz2012blood}. The multiscale approach introduced by \cite{frangi1998multiscale} develops a vessel enhancement filter with the analysis of multiscale second order local structure of an image (Hessian), and obtains a vesselness measure by using the eigenvalues of the Hessian. The matched filtering method described by \cite{hoover2000locating} employs different threshold probes to extract blood vessels from matched filter response images. In \cite{quek2001vessel}, the methodology based on vessel tracking applies a wave propagation and traceback mechanism to label each pixel the likelihood of belonging to vessels in angiography images. The mathematical morphology with the extraction of vessel centerlines \cite{mendonca2006segmentation} is also developed to find the morphological characteristics of retinal blood vessels. Model based methods generally use geometric deformable models \cite{sum2008vessel}, parametric deformable models \cite{al2009active}, vessel profile models \cite{lam2010general} and active contour models \cite{zhao2015automated} for blood vessel segmentation.

\begin{figure}
  \centering
  \subfigure[]{
	\includegraphics[width=0.95in]{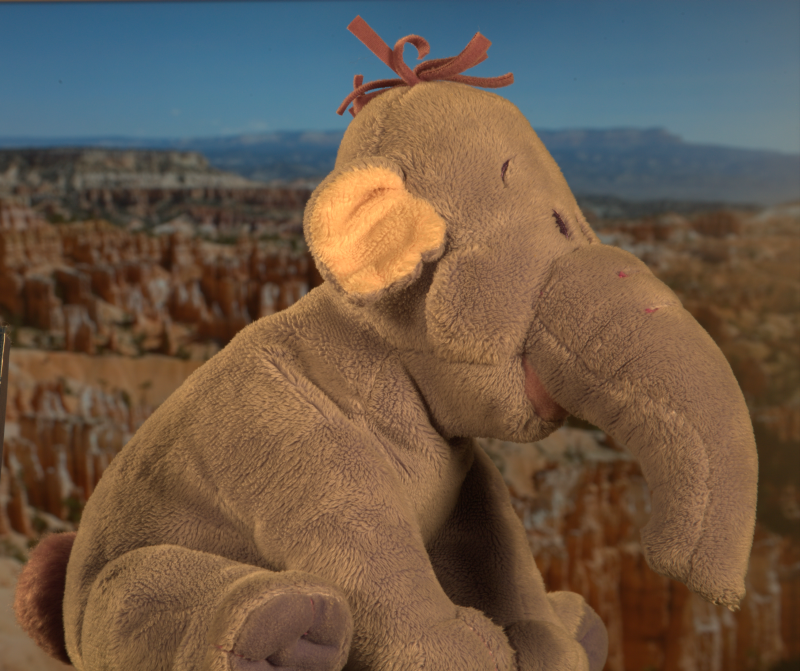}
  }
  \subfigure[]{
	\includegraphics[width=0.95in]{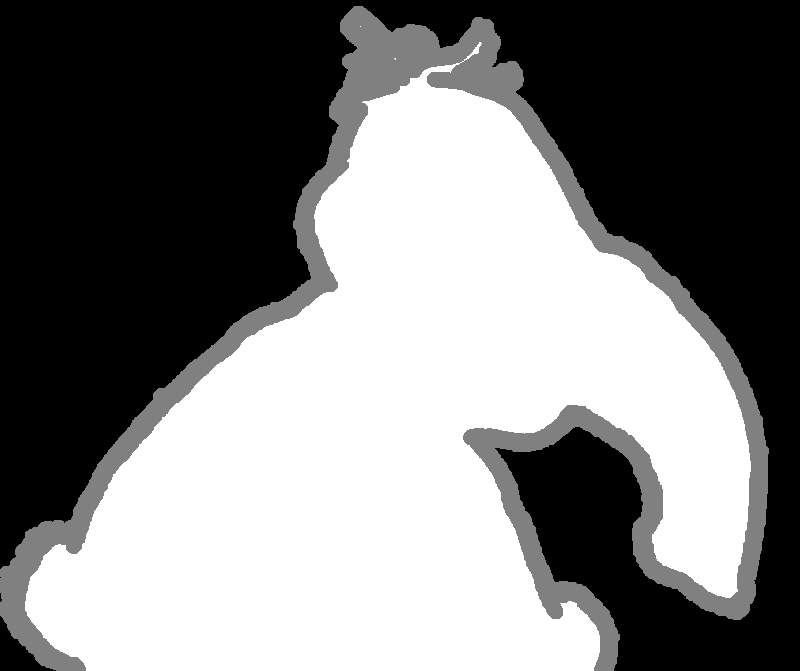}
  }
  \subfigure[]{
	\includegraphics[width=0.95in]{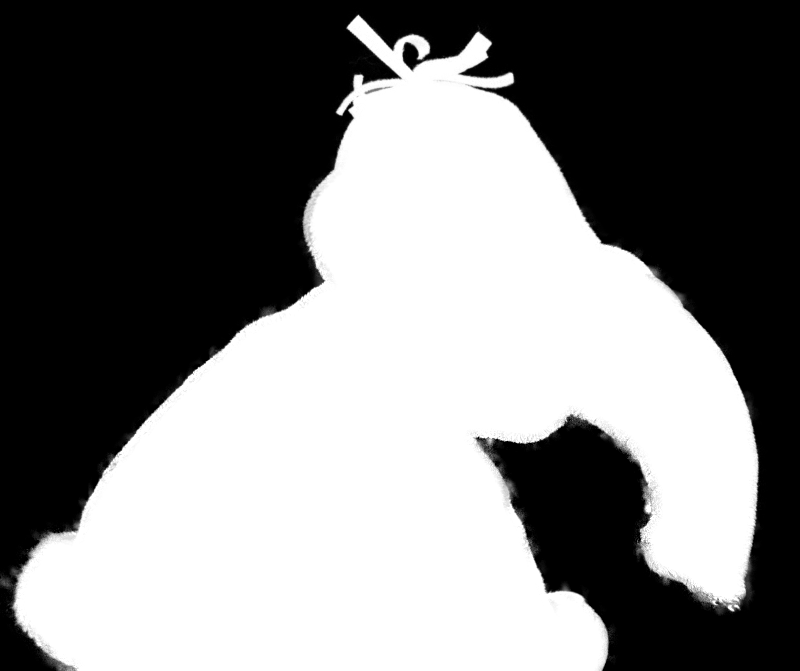}
  }
  \caption{ The process of image matting. (a) An input image. (b) A trimap generated by the user. The white, black and gray regions belong to the foreground, background and unknown regions, respectively. (c) A result achieved by \cite{zheng2009learning}.}
  \label{mattingTrimap}
\end{figure}

Image matting refers to the problem of accurately extracting a foreground object from an input image. Generally image matting includes two main steps. The first step is generating a user specified \emph{trimap}. An example of a user specified trimap is shown in Fig.\ref{mattingTrimap}(b). The trimap is a hand-drawn segmented image, which separates the input image into three regions: foreground (shown in white), background (shown in black) and unknown (shown in gray). The second step is applying the image matting model to extract the pixels belonging to the foreground object from the unknown regions, based on the samples of foreground and background pixels marked by the user. An exemplary result achieved by \cite{zheng2009learning} is shown in Fig.\ref{mattingTrimap}(c). Image matting is very useful in many important applications, such as image (or video) segmentation, image editing, video production, new view synthesis, and film making. To the best of our knowledge, image matting has never been employed before to extract blood vessels from fundus images. The major reason is that for retinal blood vessel segmentation, generating a user specified trimap is a tedious and time-consuming task. In other words, it is not appropriate to create a trimap manually for retinal blood vessel segmentation. In addition, a normal image matting model needs to be designed carefully to improve the performance of blood vessel segmentation. In order to overcome these problems, region features of blood vessels are applied to generate the trimap automatically. Then a hierarchical image matting model is proposed to extract the pixels of blood vessel from the unknown regions. More specifically, a hierarchical strategy utilizing the continuity and extendibility of retinal blood vessels is integrated into the image matting model for blood vessel segmentation. The proposed model is evaluated on three public available datasets DRIVE, STARE, and CHASE\_DB1, which have been widely used by other researchers to develop their own methods. The vessel segmentation performance demonstrates the efficiency and effectiveness of the proposed hierarchical image matting model.

The rest of this paper is structured as follows: Section II provides some background knowledge of image matting, and a brief introduction of vessel enhancement filters applied in our work. Section III details the process of generating the trimap of a fundus image automatically, and the proposed hierarchical image matting model. Section IV introduces the public available datasets and the commonly used evaluation metrics. Section V presents the experimental results. In Section VI, the discussion and conclusion are given.

\section{Related Work}
In this section, we will first review some background knowledge of image matting, and then briefly introduce vessel enhancement filters used in our work.

\subsection{Image Matting}
As aforementioned, image matting aims to accurately extract the foreground given a trimap of an image. Specifically, the input image $I(z)(z=(x,y))$ is modeled as a linear combination of a foreground image $F(z)$ and a background image $B(z)$:
\begin{equation}\label{equ:mattingModel}
  I(z) = \alpha_z F(z) + (1-\alpha_z)B(z)
\end{equation}
where $\alpha_z$, called \emph{alpha matte}, is the opacity of the foreground. $\alpha_z$ ranges from $0$ to $1$. If $\alpha_z$ is constrained to be either $0$ or $1$, then the matting problem becomes the segmentation problem, where each pixel belongs to either foreground or background.

After obtaining the user specified trimap, to infer the alpha matte in the unknown regions, Chuang \emph{et al.} \cite{chuang2001bayesian} uses sets of Gaussian distribution to model the color distributions of the foreground and background images, and estimates the optimal alpha value with a maximum-likelihood criterion. In \cite{levin2008closed}, Levin \emph{et al.} derives an effective cost function from the assumption that the foreground and background colors are locally smooth, and employs this function to find the optimal alpha matte. Zheng \emph{et al.} \cite{zheng2009learning} proposes a local learning based approach and a global learning based approach to perform image matting. In \cite{he2010fast}, Kaiming \emph{et al.} solves a large kernel matting Laplacian, and achieves a fast matting algorithm. In \cite{shahrian2012weighted}, Shahrian and Rajan analyze the texture and color features of the image, and optimize an objective function containing the color and texture components to choose the best foreground and background pair for image matting. Shahrian \emph{et al.} \cite{shahrian2013improving} expands the sampling range of foreground and background regions, and collects a representative set of samples to estimate the alpha matte. In \cite{cho2014consistent}, Cho \emph{et al.} presents an image matting method to extract alpha mattes across sub-images of a light field image.

\subsection{Vessel Enhancement filters}
Vessel enhancement filters plays an important role in retinal blood vessel segmentation \cite{fraz2012blood}. Here two effective filters \cite{fraz2012ensemble,bankhead2012fast} used in our work are introduced.

\subsubsection{Morphologically Reconstructed Filter}
Morphologically reconstructed filter is an effective tool for blood vessel enhancement \cite{fraz2012ensemble}. For each input fundus image $I$, the green channel image $I_g$ is extracted firstly since $I_g$ has the best vessel-background contrast \cite{marin2011new}. Then the morphological top-hat transformation is performed:
\begin{equation}\label{equ:mt1}
  I_{th}^{\theta}=I_g^c-(I_g^c\circ S_e^{\theta})
\end{equation}
where $I_g^c$ is the complement image of $I_g$, $I_{th}^{\theta}$ represents the top-hat transformed image, $S_e$ is a structuring element for morphological opening $\circ$, and $\theta$ specifies the angle (in degrees) of the structuring element. The structuring element is of 1-pixel width and 21-pixels length, which approximately fits the diameter of the biggest vessels in the fundus images \cite{fraz2012ensemble}. Since the morphological top-hat transformation transformation given in Equation \eqref{equ:mt1} can only brighten blood vessels in one direction, the sum of top-hat transformation $I_{th}^{\theta}$ along each direction is performed in order to enhance the whole vessel image:
\begin{equation}\label{equ:mt2}
  I_{mr}=\sum_{\theta \in A} I_{th}^{\theta}
\end{equation}
where $I_{mr}$ represents the enhanced vessel image using morphologically reconstructed filter, "A" is the set of angles of the structuring element and defined as $\{x|0<x<\pi \; \& \; x \; mod \; (\pi/12)=0\}$.

\subsubsection{Isotropic Undecimated Wavelet Filter}
The isotropic undecimated wavelet filter has been used for blood vessel segmentation, and has a good performance on vessel enhancement \cite{bankhead2012fast}. Applied to a signal $c_0=I_g$ ($I_g$ is the input green channel image), scaling coefficients are computed by convolution with a filter $h^{\uparrow j}$ firstly:
\begin{equation}
  c_{j+1}=c_j\ast h^{\uparrow j}
\end{equation}
where $h_0=[1,4,6,4,1]/16$ is derived form the cubic B-spline, $h^{\uparrow j}$ is the upsampled filter obtained by inserting $2^j-1$ zeros between each pair of adjacent coefficients of $h_0$. Wavelet coefficients are the difference between two adjacent sets of scaling coefficients, i.e.,
\begin{equation}
  w_{j+1}=c_j-c_{j+1}
\end{equation}
Reconstruction of the original signal from all wavelet coefficients and the final set of scaling coefficients is straightforward, and requires only addition.
The final enhanced vessel image is depicted as follows:
\begin{equation}
  I_{iuw}=c_n+\sum_{j=1}^n w_j
  \label{equ:Ie2}
\end{equation}
where $I_{iuw}$ represents the enhanced vessel image using isotropic undecimated wavelet filter. In blood vessel segmentation, wavelet scales: $2-3$ are selected according to \cite{bankhead2012fast}.

\begin{figure}
  \centering
  \subfigure[]{
	\includegraphics[width=1in,height=1in]{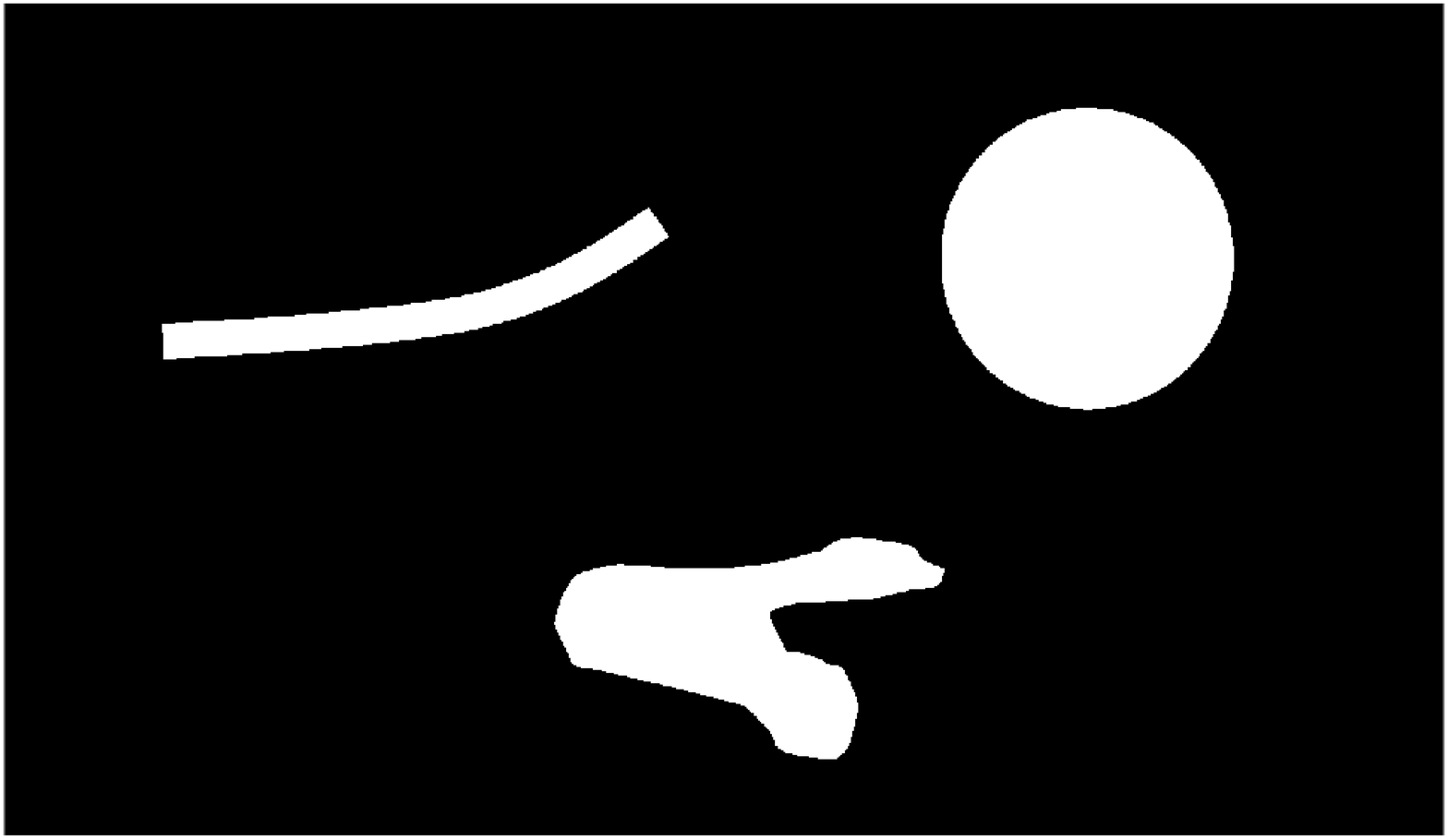}
  }
  \subfigure[]{
	\includegraphics[width=1in,height=1in]{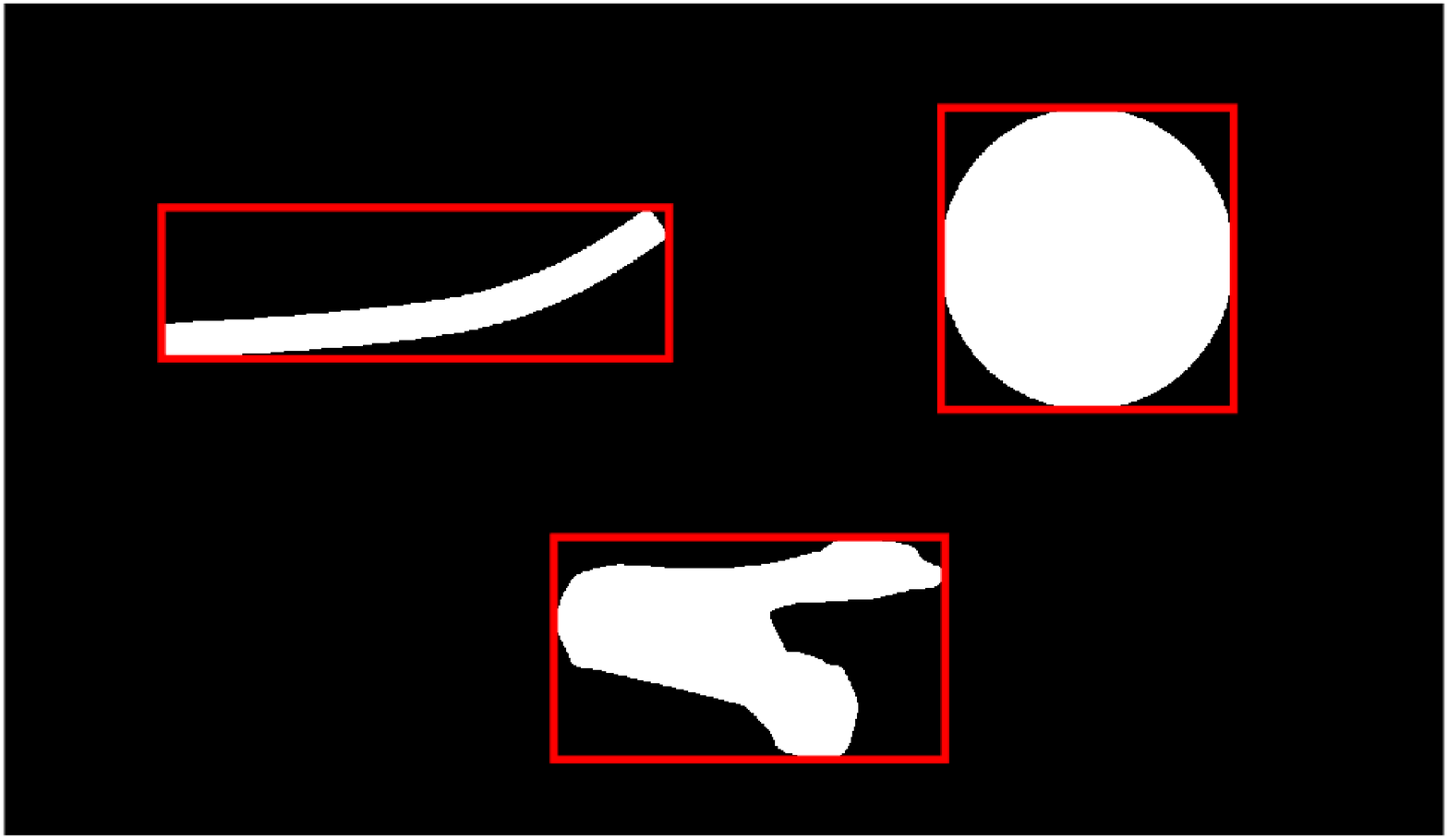}
  }
  \subfigure[]{
	\includegraphics[width=1in,height=1in]{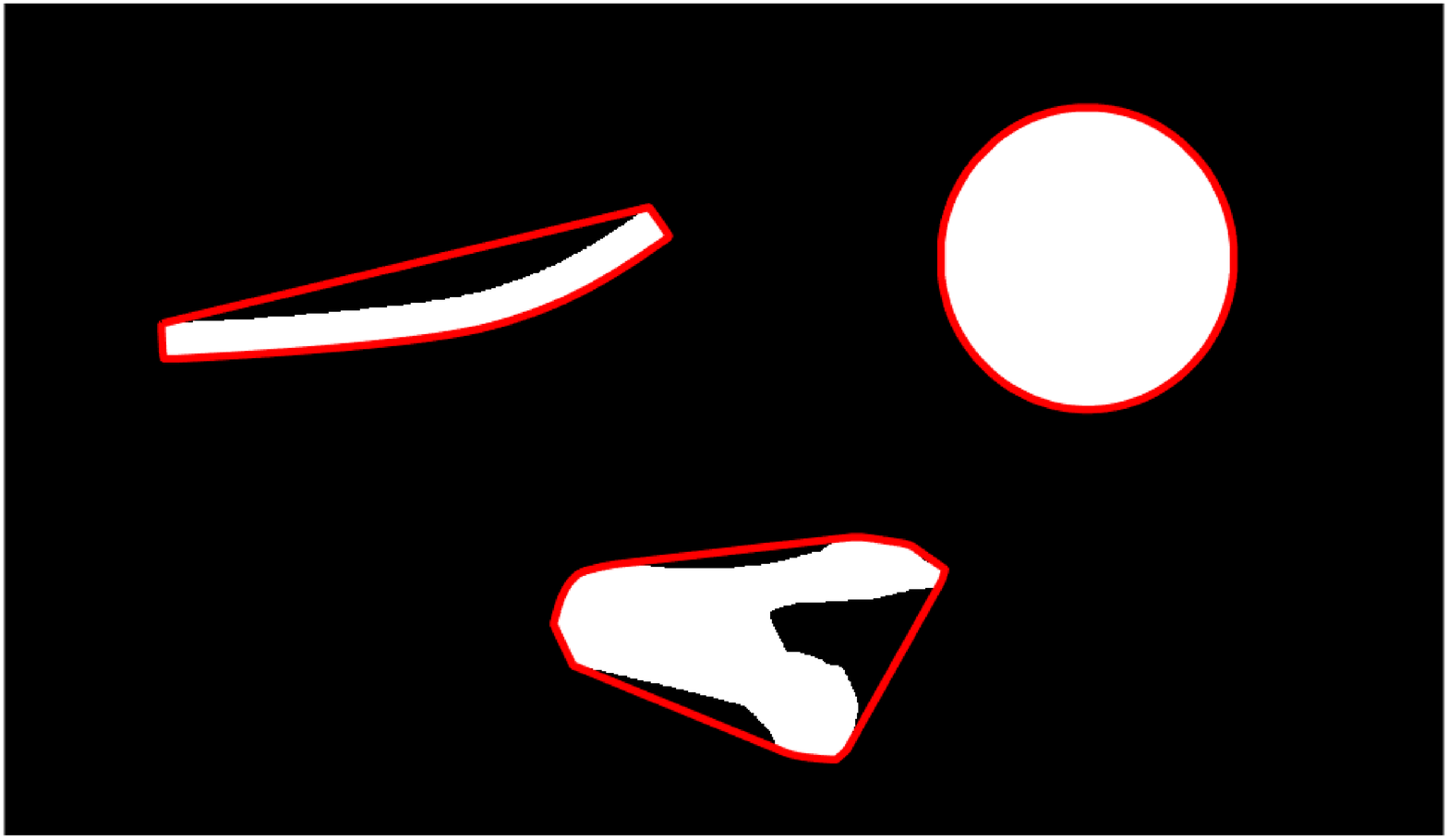}
  }
  \caption{An example to illustrate the bounding box and convex hull. (a) The exemplary image. (b) The image for the illustration of bounding box. The red boxes are the bounding boxes. (c) The image for the illustration of convex hull. The red polygons are the convex hulls.}
\label{fig:box}
\end{figure}

\section{Methodology}
In this section, the process of generating the trimap of an input fundus image automatically is introduced, followed by detailing the proposed hierarchical image matting model.

\begin{table}[htbp]
  \centering
  \caption{The default threshold values of region features:\emph{Extent}, \emph{VRatio}, \emph{Solidity} and their recommended ranges used in this work}
    \begin{tabular}{|c|cccc|}
    \hline
    Threshold values & $e_1$    & $e_2$    & $r$     & $s$ \\
    \hline
    Default values & $0.35$  & $0.25$  & $2.2$   & $0.53$ \\
    \hline
    Recommended Ranges & $[0.2,0.4]$ & $[0.15,0.3]$ & $[2,6]$ & $[0.4,0.6]$ \\
    \hline
    \end{tabular}%
  \label{tab:parametersTable}%
\end{table}%

\subsection{Trimap Generation}
Region features of blood vessels have been used for blood vessel segmentation and performed well on segmentation accuracy and computational efficiency \cite{fan2016automated}. In this paper, region features of blood vessels are applied to generate the trimap of the input fundus image automatically. The definitions of regions features are given as follows:
\begin{itemize} \itemsep -1pt
\item \textbf{\emph{Area}} indicates the actual number of pixels in the region.

\item \textbf{\emph{Bounding Box}} specifies the smallest rectangle containing the region. An example for the illustration of bounding box is shown in Fig.\ref{fig:box}(b).

\item \textbf{\emph{Extent}} is the ratio of pixels in the region to pixels in the total bounding box.

\item \textbf{\emph{VRatio}} is the ratio of the length to the width of the bounding box.

\item \textbf{\emph{Convex Hull}} specifies the smallest convex polygon that can contain the region. An example for the illustration of convex hull is shown in Fig.\ref{fig:box}(c).

\item \textbf{\emph{Solidity}} is the ratio of pixels in the region to pixels in the total convex hull.
\end{itemize}
The default threshold values of region features:\emph{Extent}, \emph{VRatio}, \emph{Solidity} and their recommended ranges used in this work are reported in Table \ref{tab:parametersTable}. $e_1$ and $e_2$ are two threshold values of \emph{Extent} features used in this work; $r$ is the threshold value of $VRatio$ feature; $s$ is the threshold value of $Solidity$ feature. For $Area$ feature, two threshold values: $a_1=f_i\times2$ and $a_2=f_i\times35$ are used in this work. $f_i$, called the internal factor, is calculated as $d\times \frac{max(h,w)}{min(h,w)}$, where $d=21$ is approximately the diameter of the biggest vessels in fundus images\cite{fraz2012ensemble}, $h$ and $w$ are the height and width of the fundus image.

The proposed model is not sensitive to above mentioned region features. In other words, these region features can be selected in a relatively large range without sacrificing the performance. In Section V(D)-"Sensitivity analysis of threshold values of region features", empirical study is conducted to verify the insensitivity of the proposed model to the threshold values of region features.

\begin{figure}
  \centering
	\includegraphics[width=3.43in]{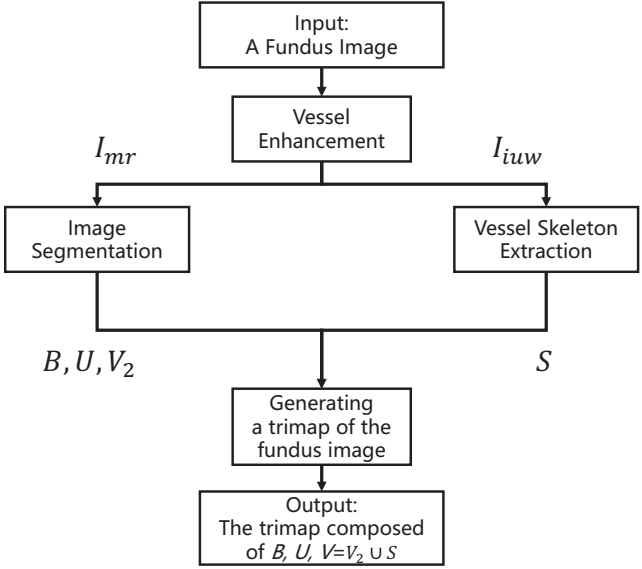}
  \caption{The process of trimap generation. \emph{B} represents the background regions; \emph{U} represents the unknown regions; $V_2$ represents the denoised preliminary vessel regions; $S$ represents the skeleton of blood vessels; $V$ represents the vessel regions.}
  \label{fig:TrimapGeneration}
\end{figure}

Creating the trimap of the input fundus image includes two main steps: 1) Image Segmentation and 2) Vessel Skeleton Extraction. The process of trimap generation is given in Fig.\ref{fig:TrimapGeneration}.

\begin{figure}
  \centering
  \subfigure[]{
	\includegraphics[width=0.95in]{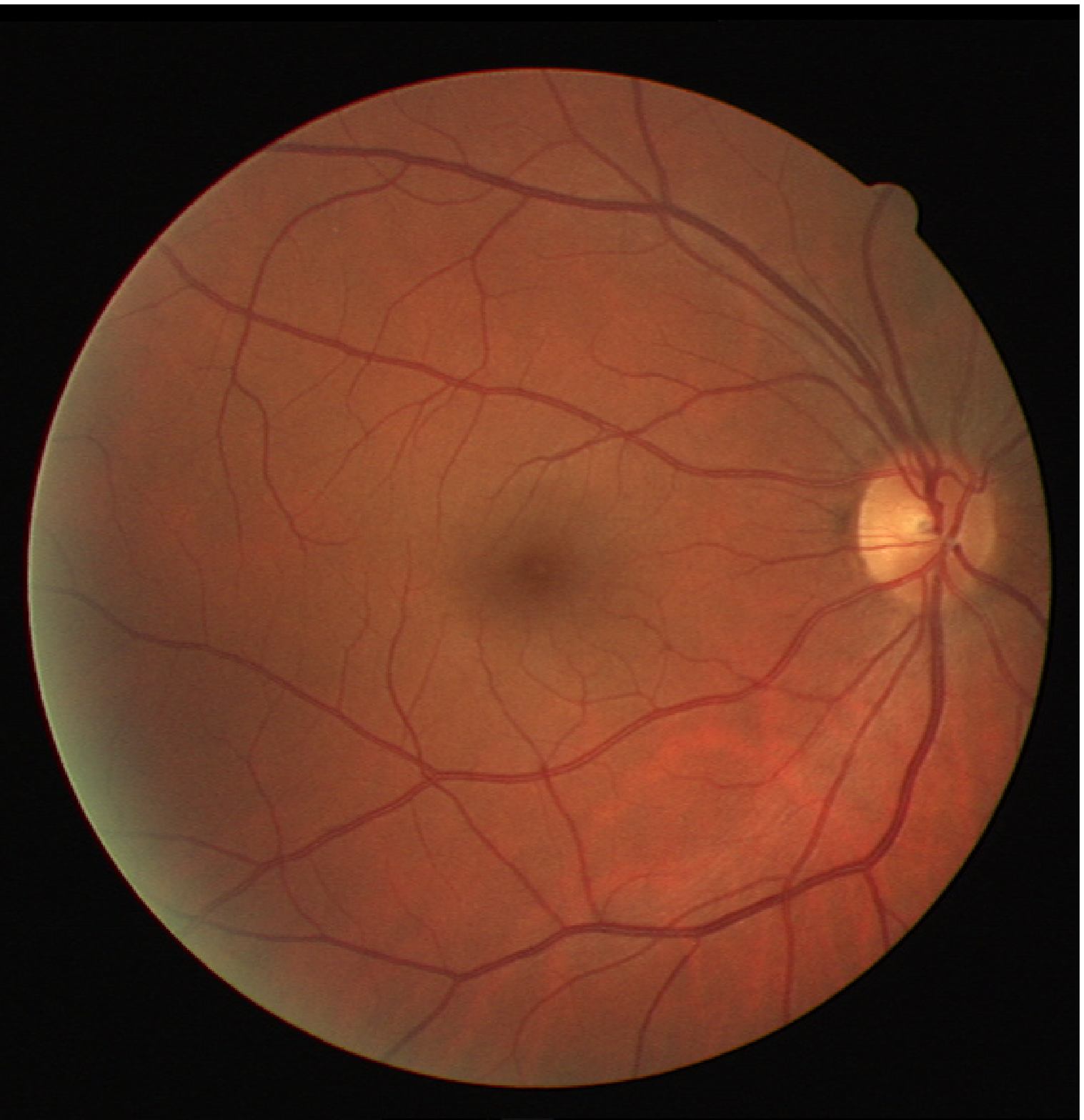}
  }
  \subfigure[]{
	\includegraphics[width=0.95in]{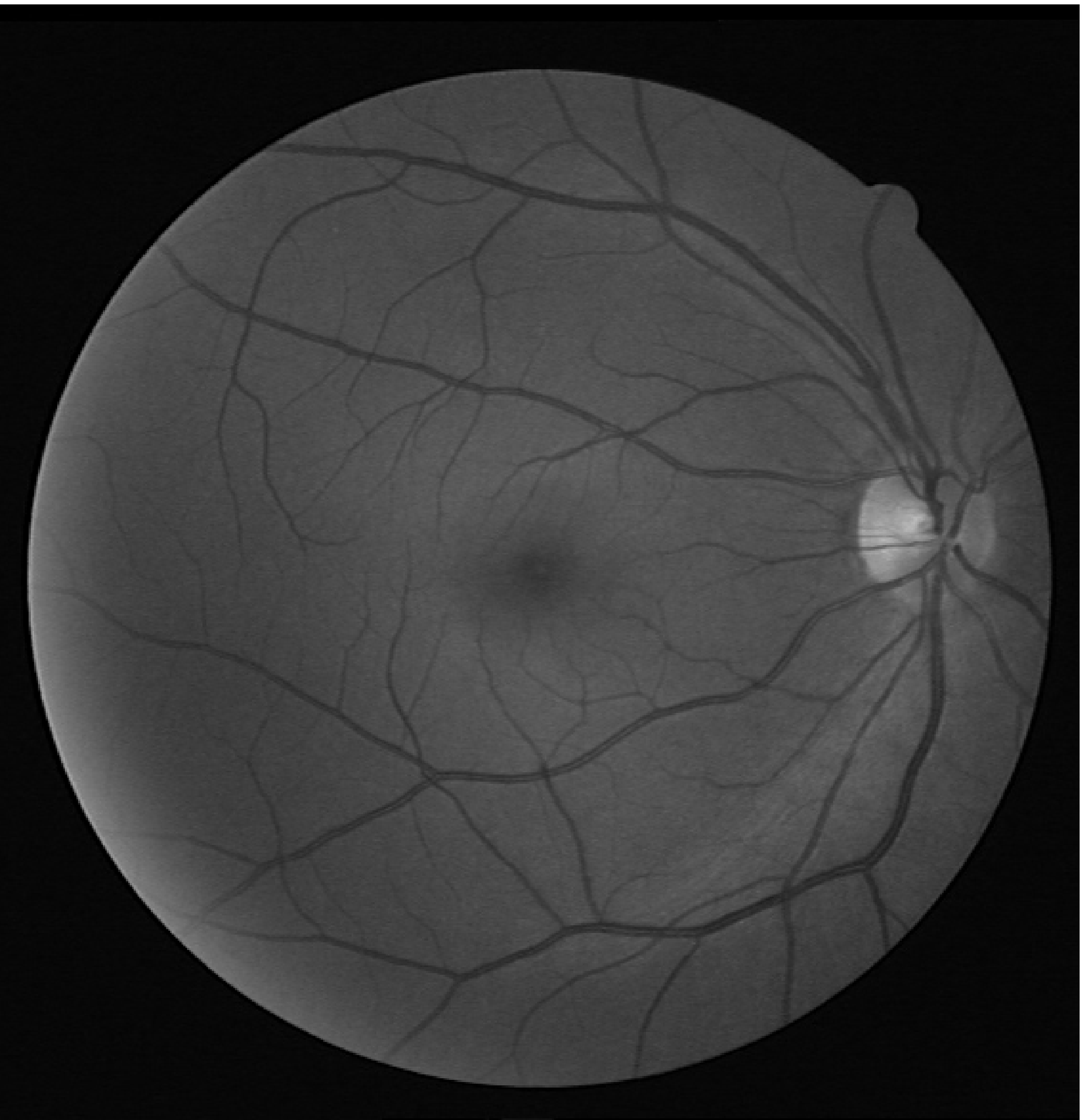}
  }
  \subfigure[]{
	\includegraphics[width=0.95in]{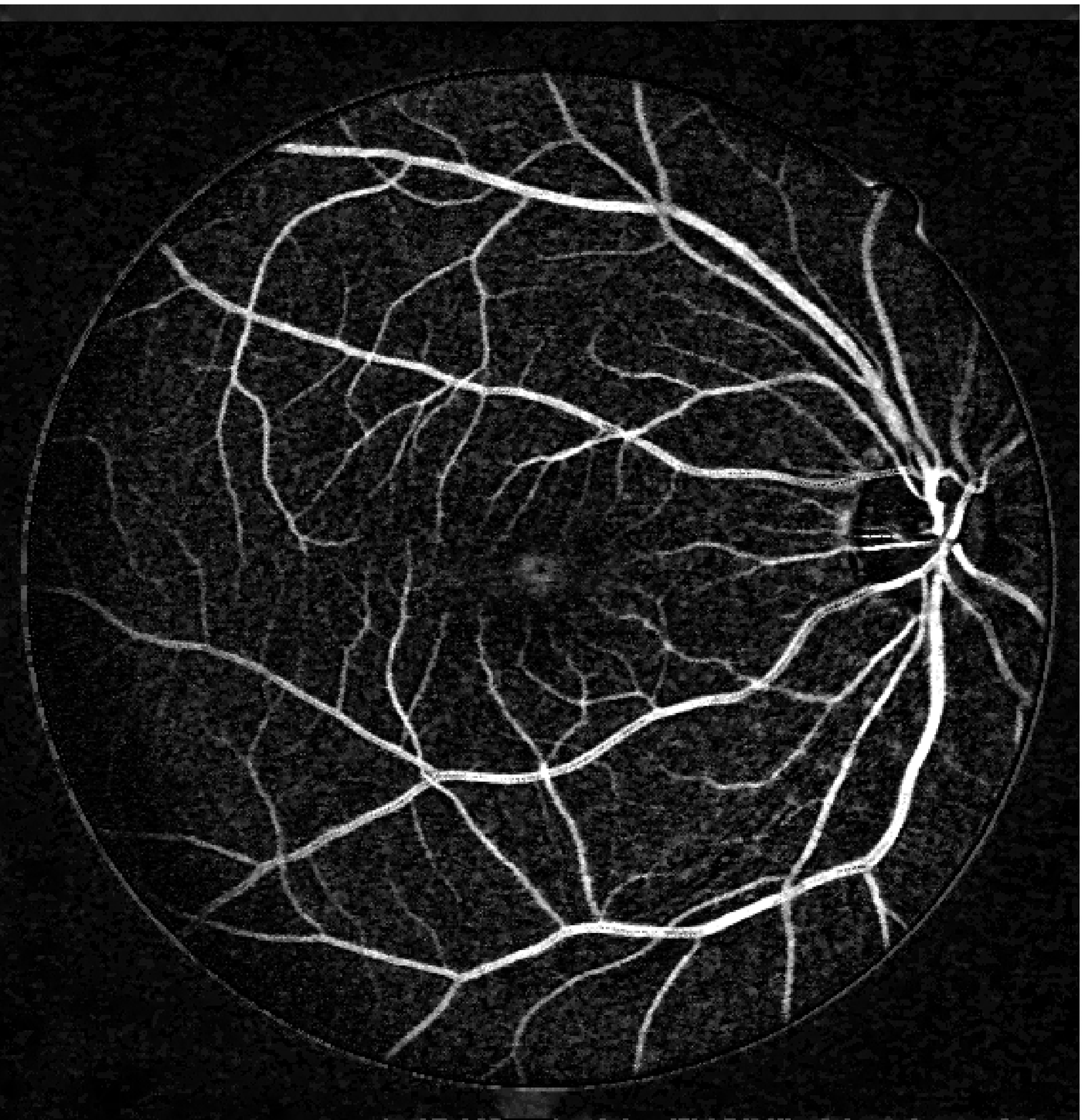}
  }
  \subfigure[]{
	\includegraphics[width=0.95in]{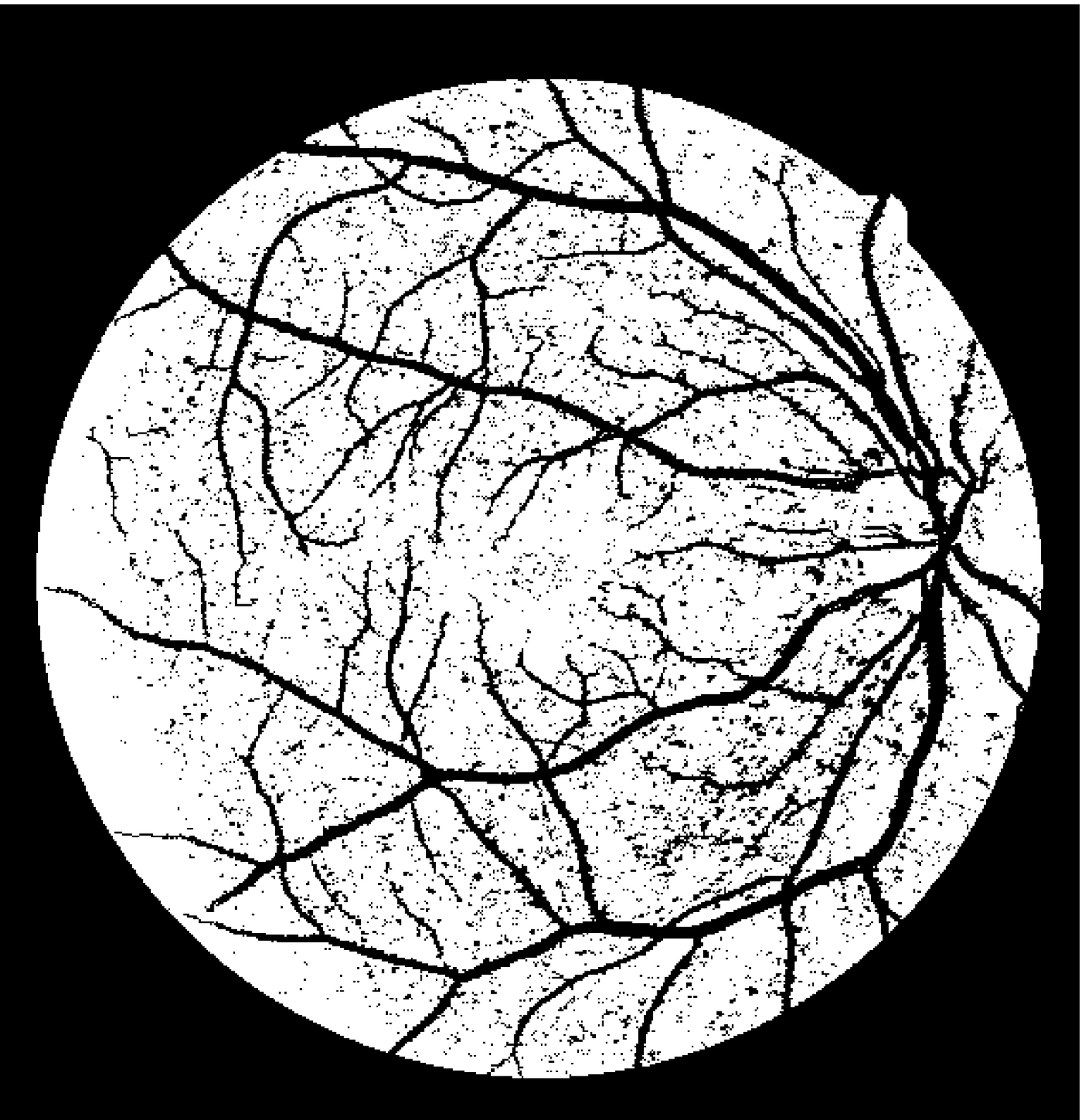}
  }
  \subfigure[]{
	\includegraphics[width=0.95in]{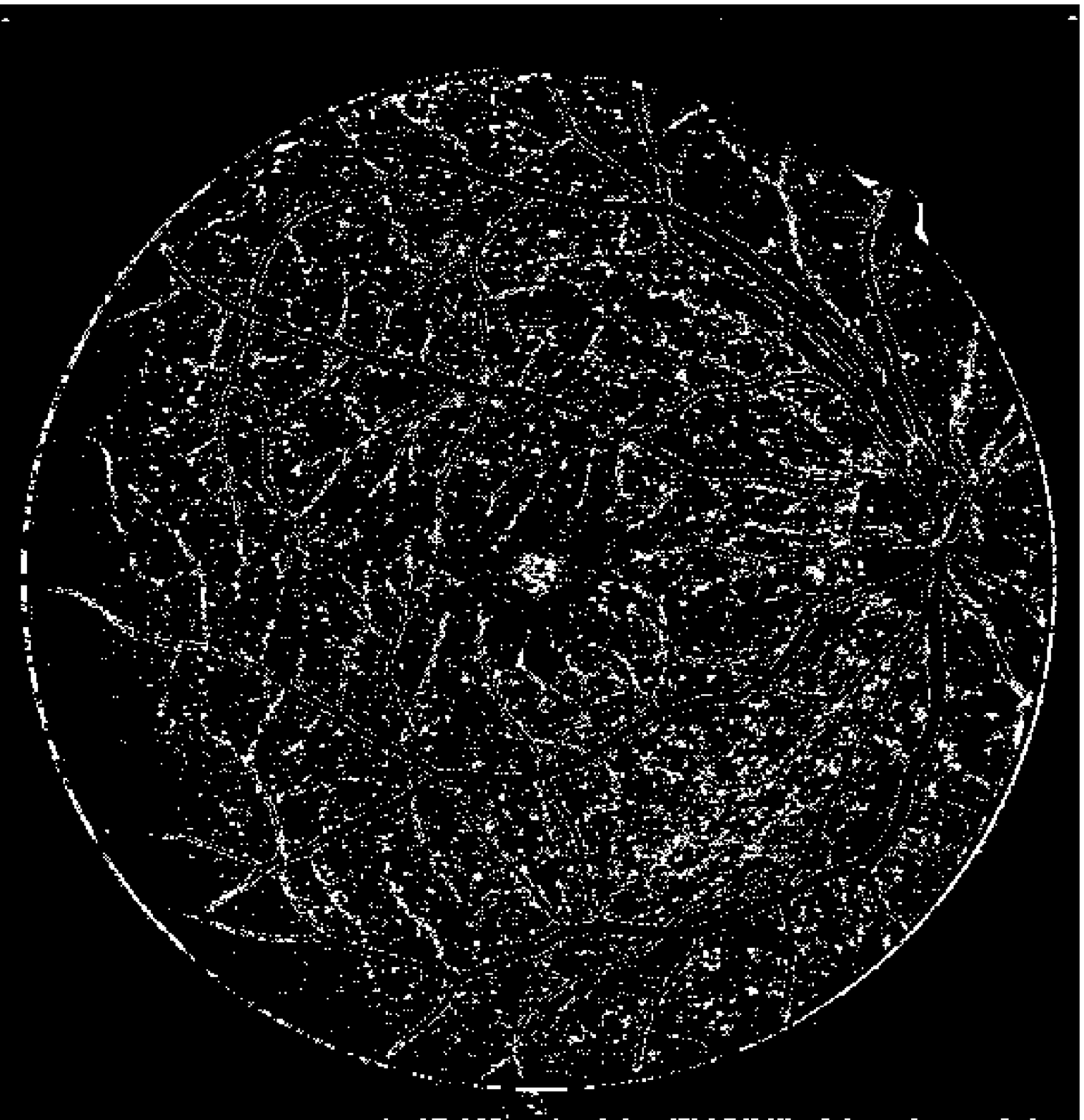}
  }
  \subfigure[]{
	\includegraphics[width=0.95in]{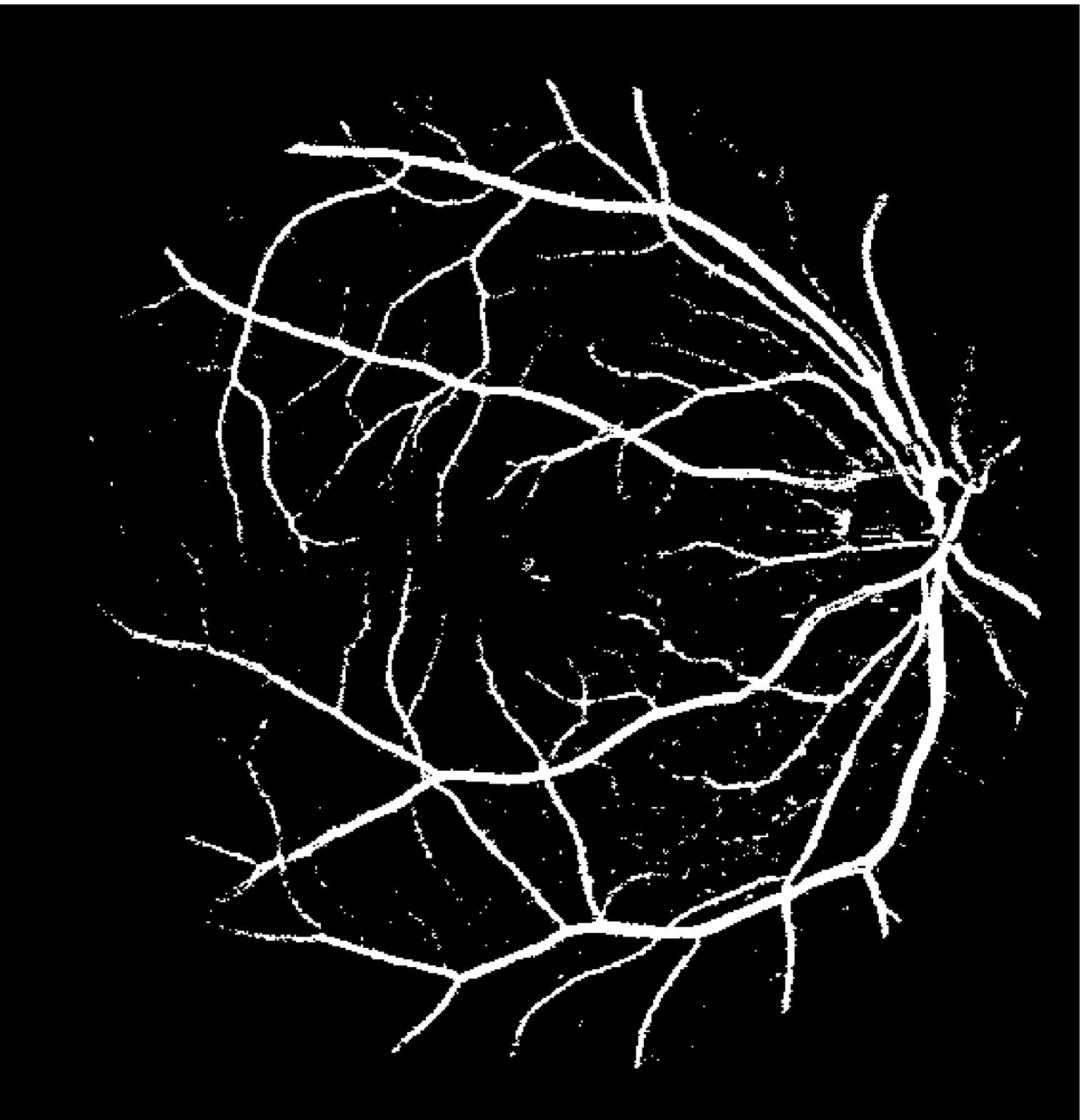}
  }
  \caption{Image segmentation.  (a) The fundus image $I$. (b) The green channel of the fundus image $I_g$. (c) The enhanced vessel image $I_{mr}$. (d) The background regions $B$. (e) The unknown regions $U$. (f) The denoised preliminary vessel regions $V_2$}
\label{fig:imageSegmentation}
\end{figure}

\subsubsection{Image Segmentation}
The goal of image segmentation is to divide the input image into three regions: the vessel (foreground), background and unknown regions. Firstly the enhanced vessel image $I_{mr}$ is segmented into three regions: the background regions ($B$), unknown regions ($U$) and preliminary vessel regions ($V_1$)
\begin{equation}
           I_{mr}= \left\{
                \begin{array}{rl}
                 B & \text{if $0<I_{mr}<p_1$} \\
                 U & \text{if $p_1\leqslant I_{mr}<p_2$} \\
                 V_1 & \text{if $p_2\leqslant I_{mr}$}
                 \end{array} \right.
\end{equation}
where $p_1=0.2$ and $p_2=0.35$ restrict the unknown region as thin as possible in order to achieve the better matting result \cite{wang2005iterative,shahrian2012weighted}. In order to remove the noise regions in $V_1$, the regions with $Area>a_1$ in $V_1$ are extracted firstly ($V_1^*$). Then regions in $V_1^*$ whose $Extent\leq e_1$ \&\& $VRatio\leq r$ \&\& $Solidity\geq s$ are abandoned, resulting in the denoised preliminary vessel
regions $V_2$. An example of image segmentation is shown in Fig.\ref{fig:imageSegmentation}.

\begin{figure}
  \centering
  \subfigure[]{
	\includegraphics[width=0.95in]{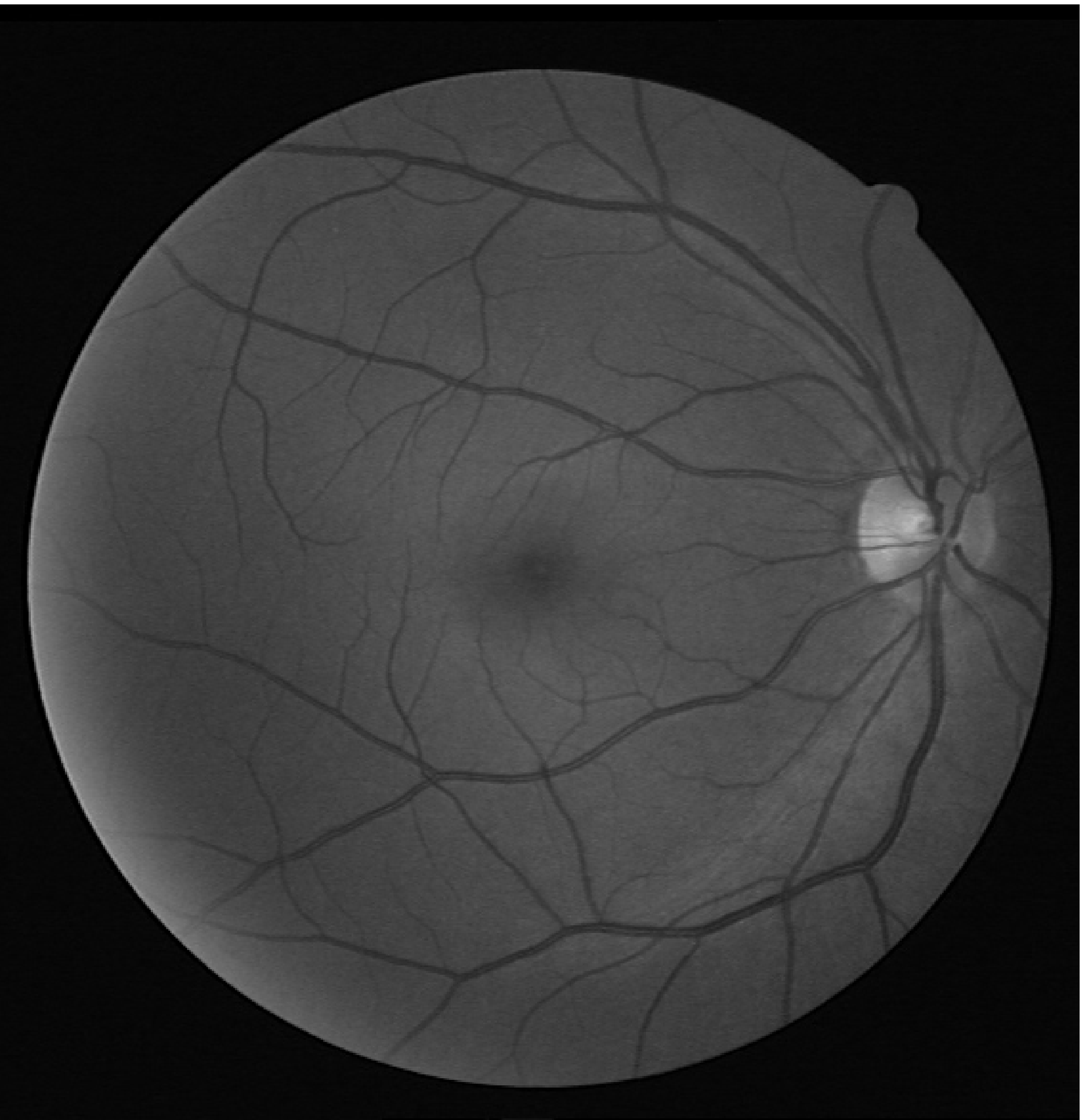}
  }
  \subfigure[]{
	\includegraphics[width=0.95in]{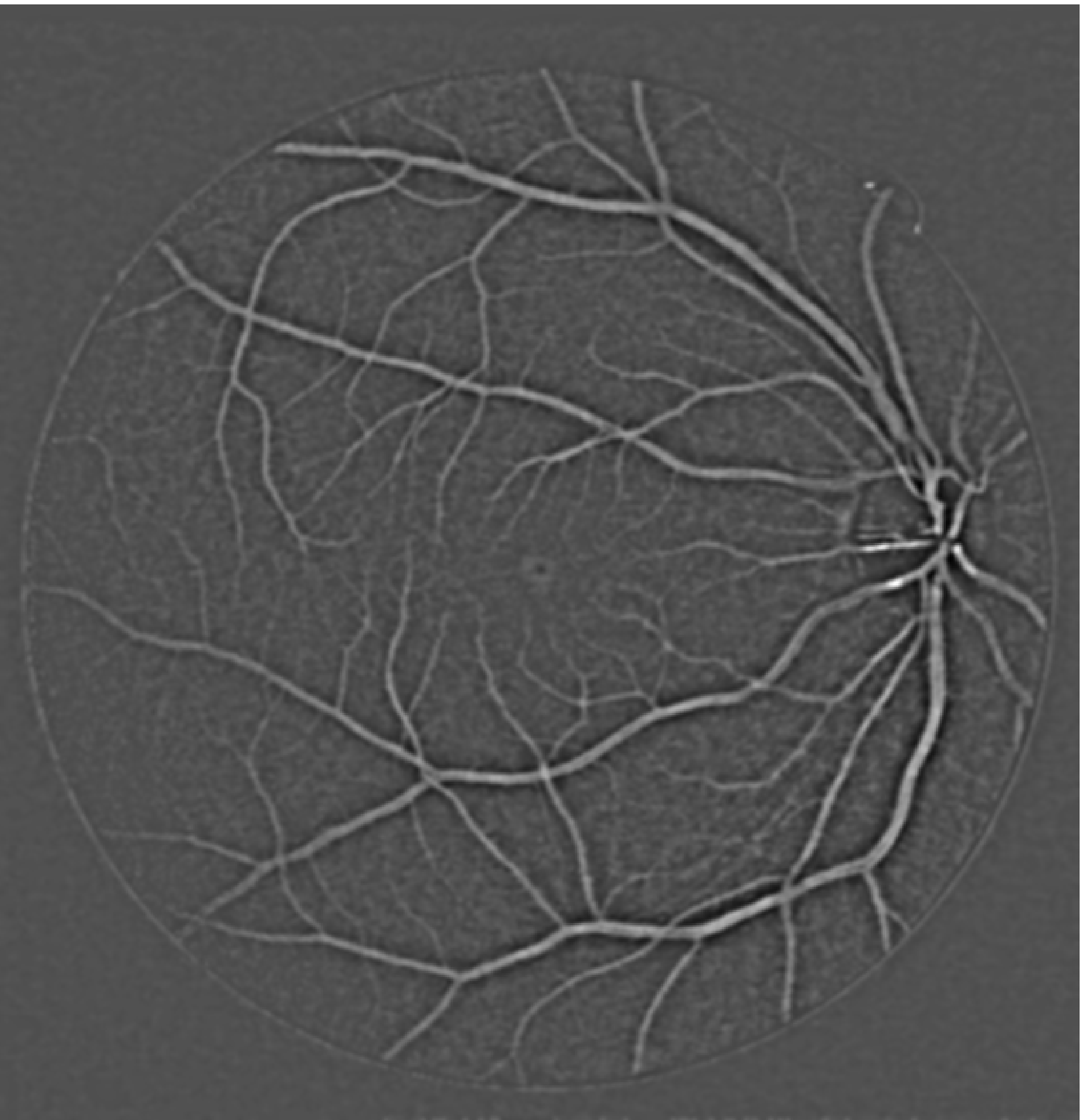}
  }
  \subfigure[]{
	\includegraphics[width=0.95in]{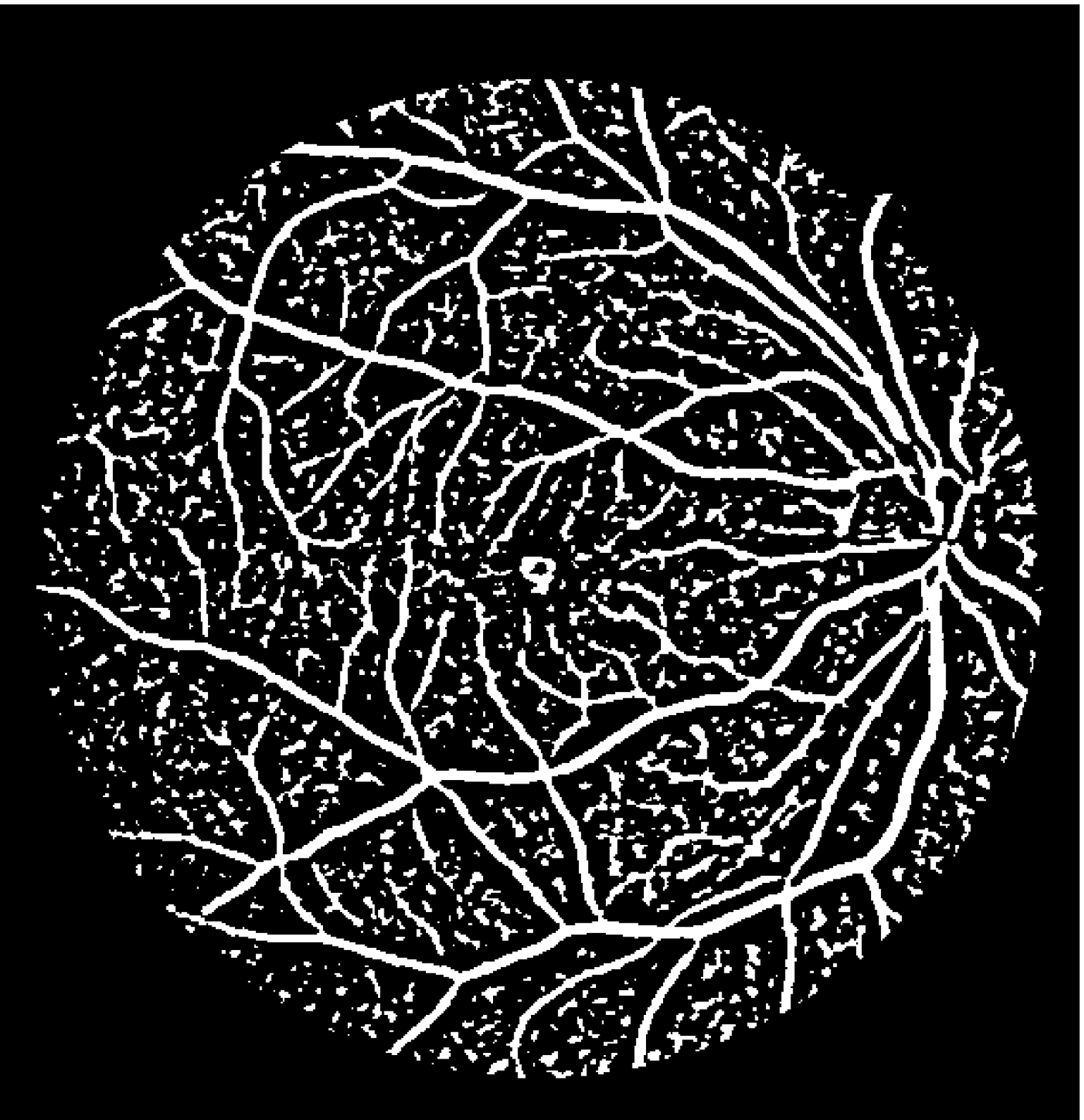}
  }
  \subfigure[]{
	\includegraphics[width=0.95in]{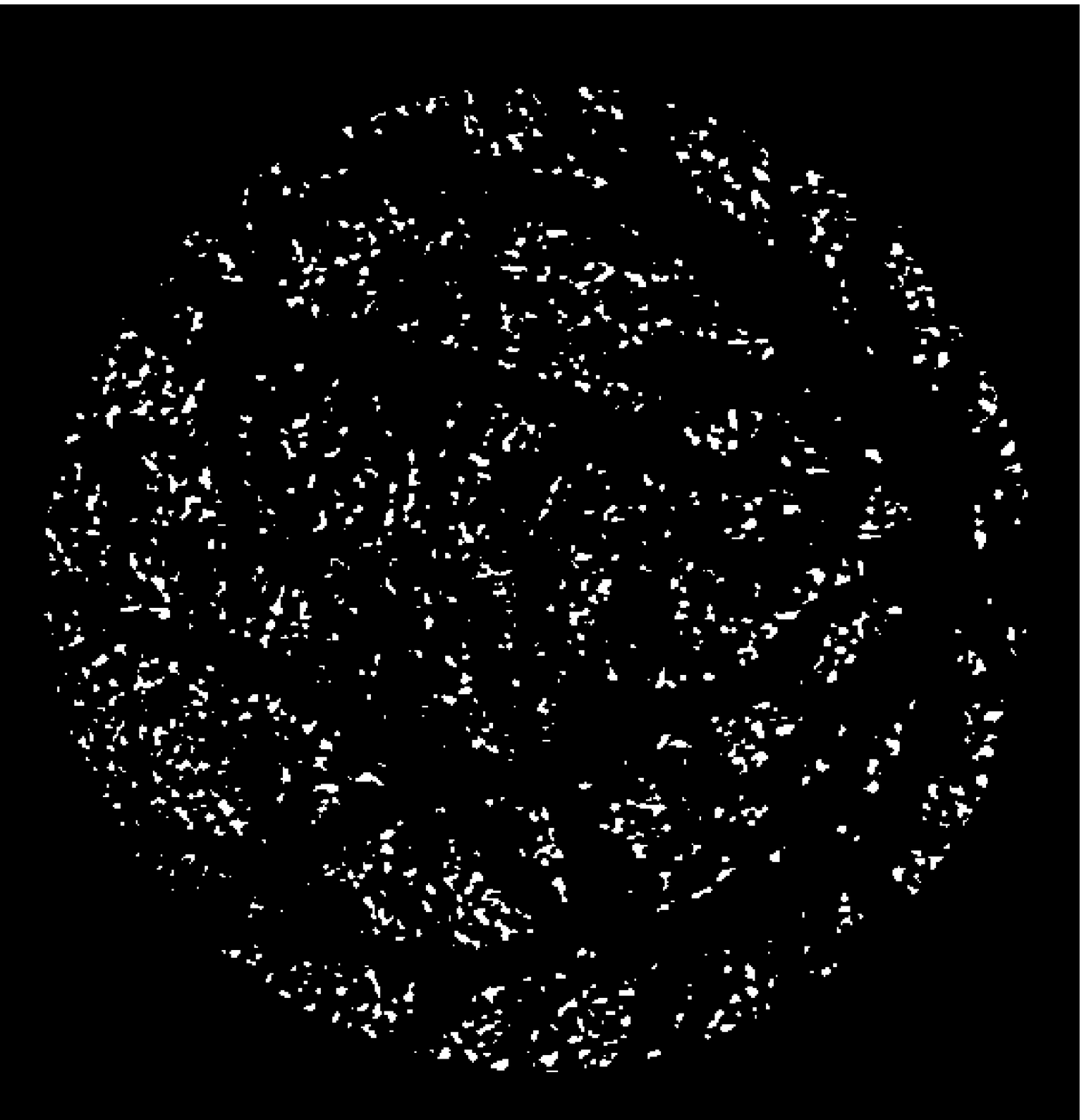}
  }
  \subfigure[]{
	\includegraphics[width=0.95in]{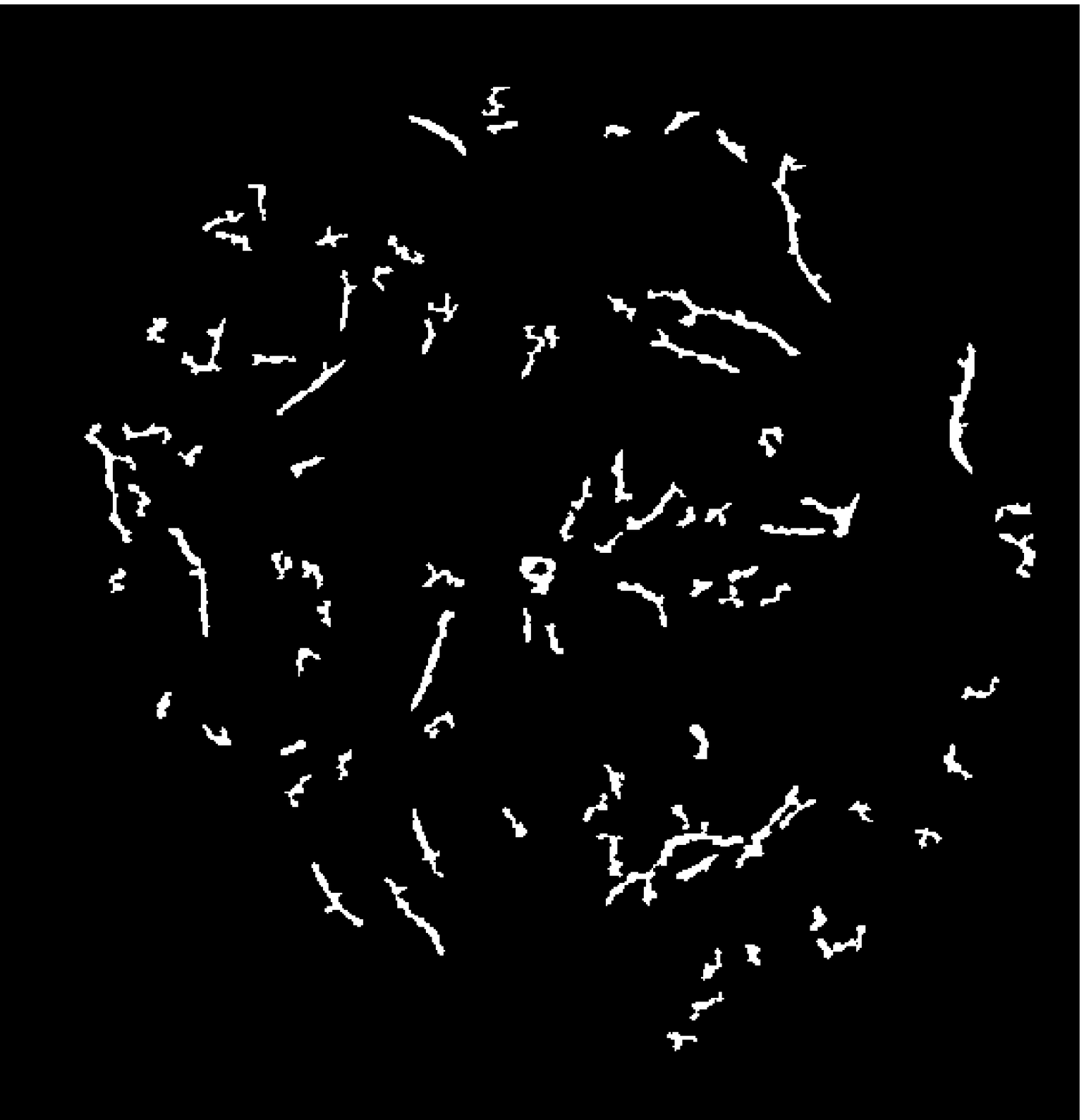}
  }
  \subfigure[]{
	\includegraphics[width=0.95in]{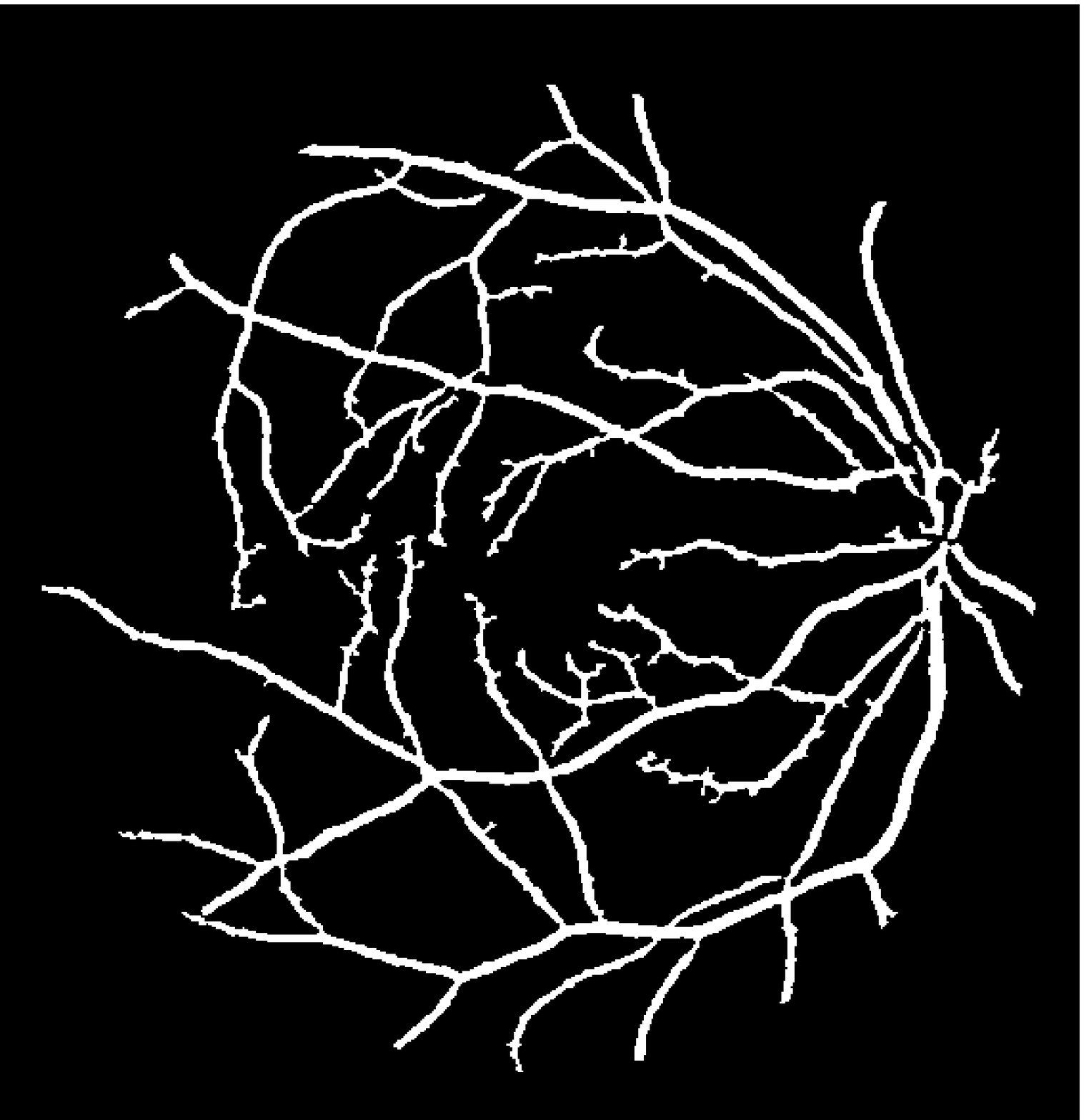}
  }
  \subfigure[]{
	\includegraphics[width=0.95in]{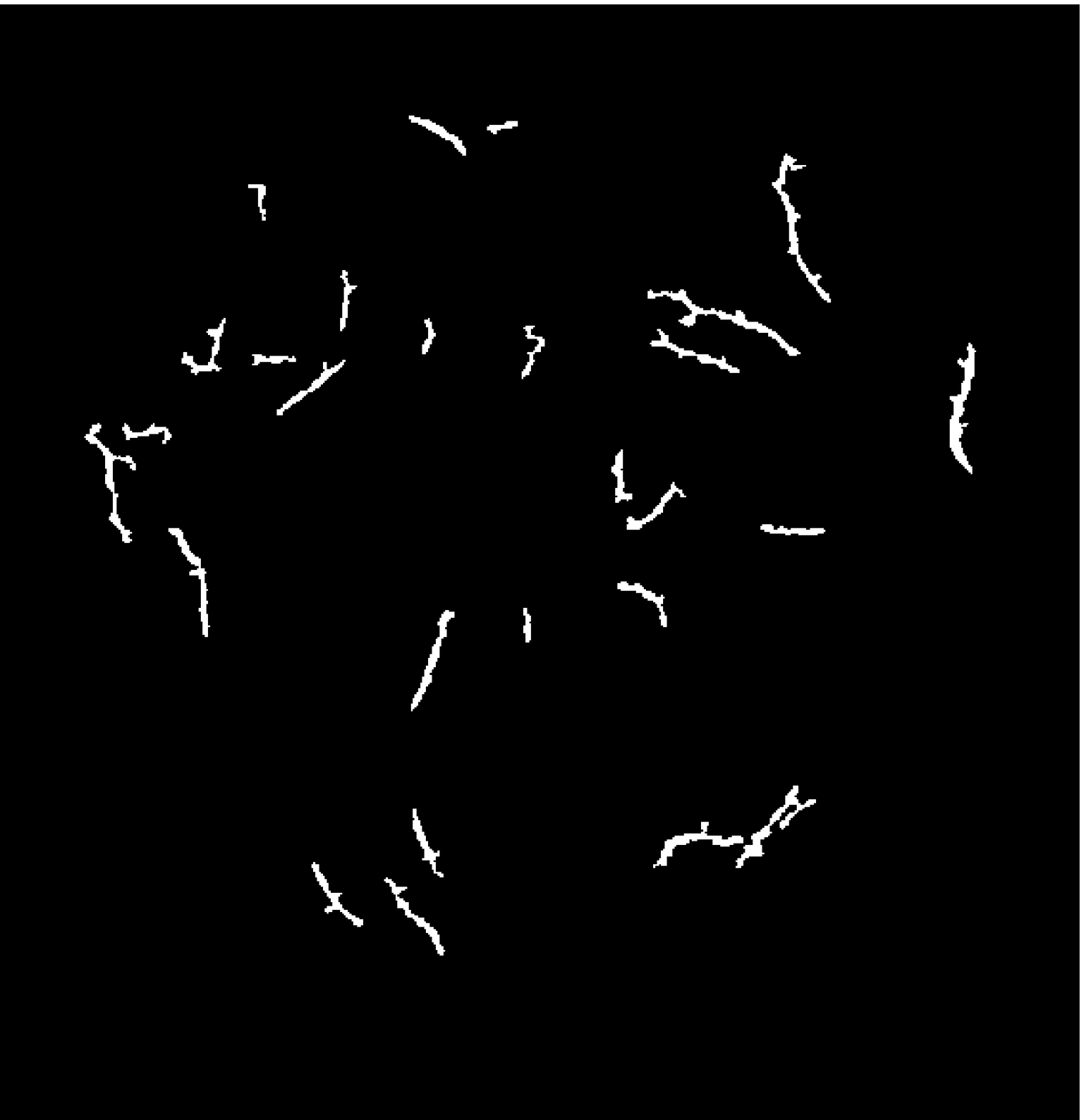}
  }
  \subfigure[]{
	\includegraphics[width=0.95in]{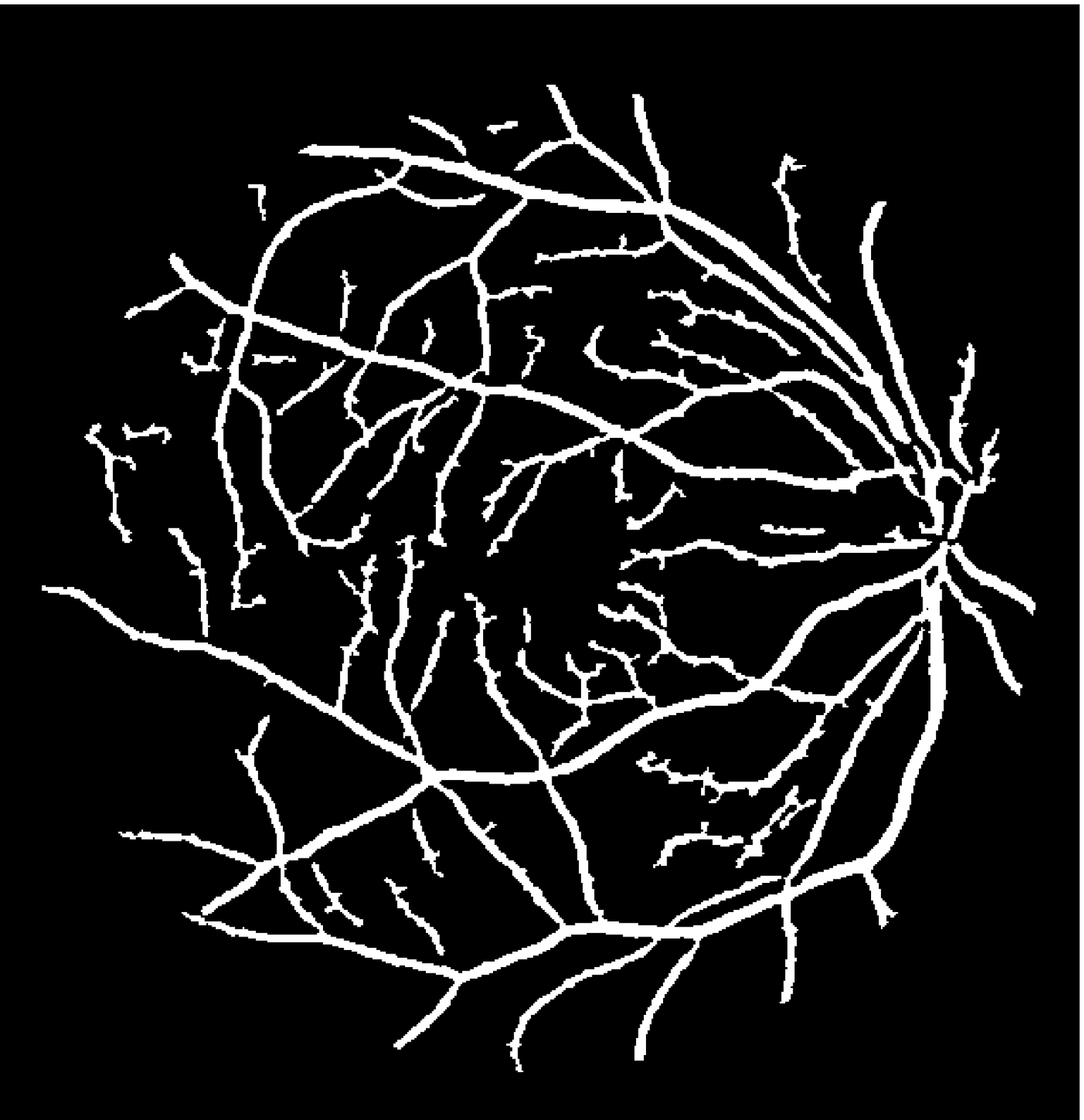}
  }
  \subfigure[]{
	\includegraphics[width=0.95in]{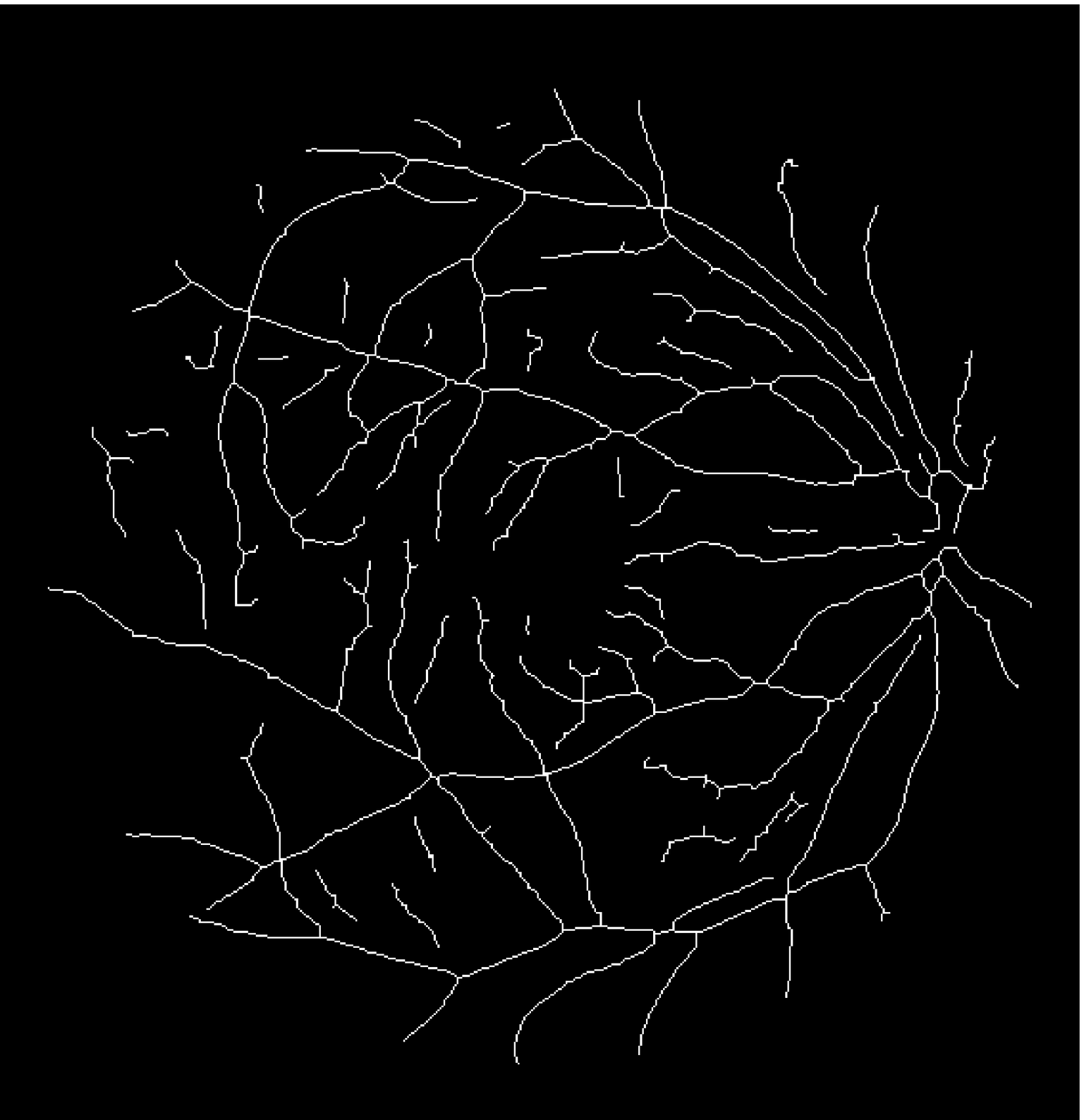}
  }

  \caption{Vessel skeleton extraction. (a) The green channel of the fundus image $I_g$. (b) The vessel enhanced image $I_{iuw}$. (c) The binary image $T$. (d) The background regions $T_1$. (e) The candidate regions $T_2$. (f) The vessel regions $T_3$. (g) $T_4$: The preserved regions in $T_2$. (h) The combined regions of $T_3$ and $T_4$. (i) The vessel skeletons $S$.}
  \label{fig:vesselSkeleton}
\end{figure}

\subsubsection{Vessel Skeleton Extraction}
Vessel Skeleton Extraction aims to further distinguish the unknown regions and provide more information on blood vessels. In Section V(B)-"Vessel Segmentation Performance", the effectiveness of vessel skeleton extraction will be presented. Firstly, a binary image $T$ is obtained by global thresholding the enhanced vessel image $I_{iuw}$.
\begin{equation}
           T = \left\{
                \begin{array}{rl}
                 1 & I_{iuw}>t \\
                 0 & I_{iuw}\leqslant t
                 \end{array} \right.
\end{equation}

\begin{figure*}\label{fig:initialize}
  \centering
  \subfigure[]{
  \includegraphics[width=2.25in,height=1.5in]{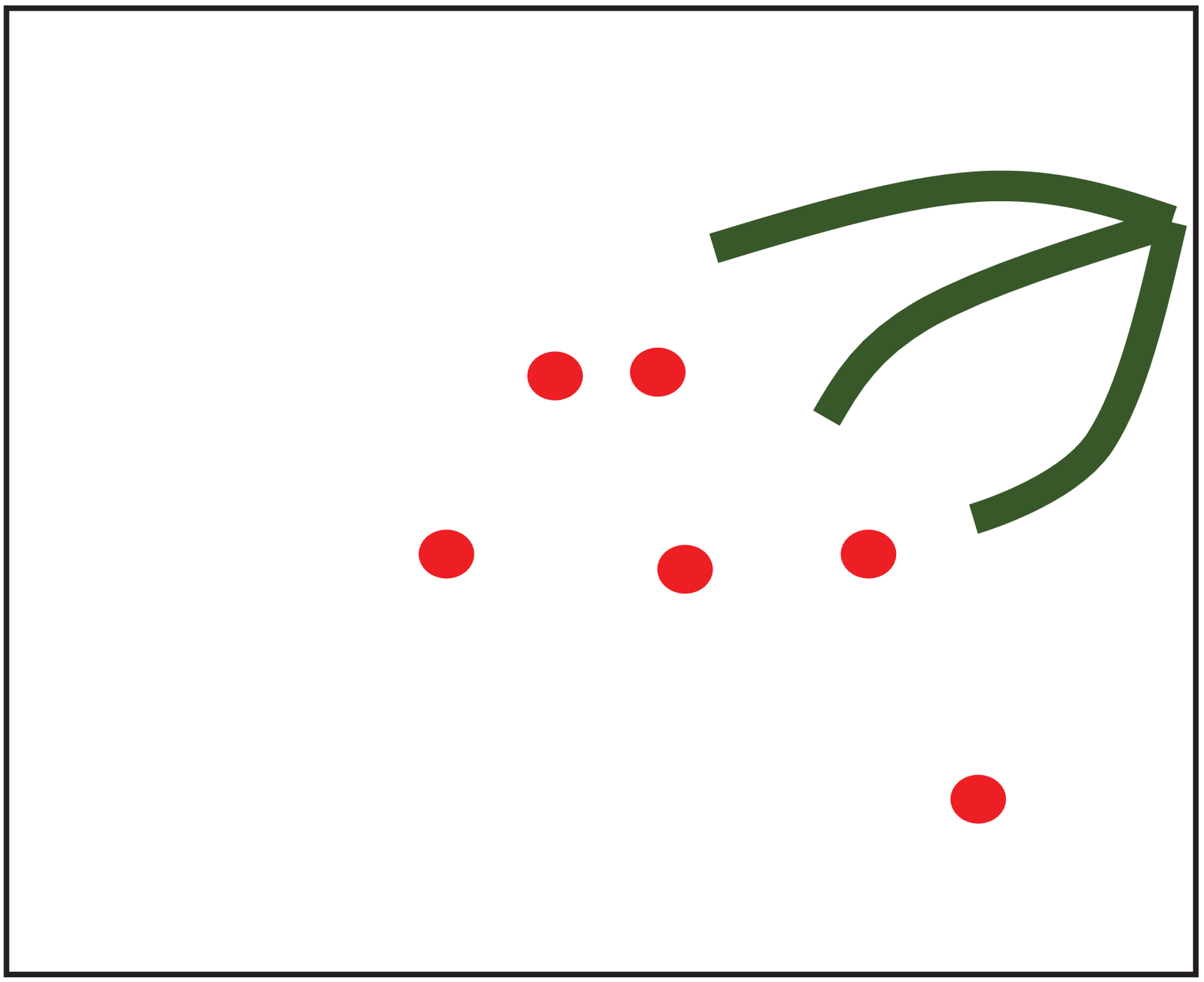}
  }
  \subfigure[]{
  \includegraphics[width=2.25in,height=1.5in]{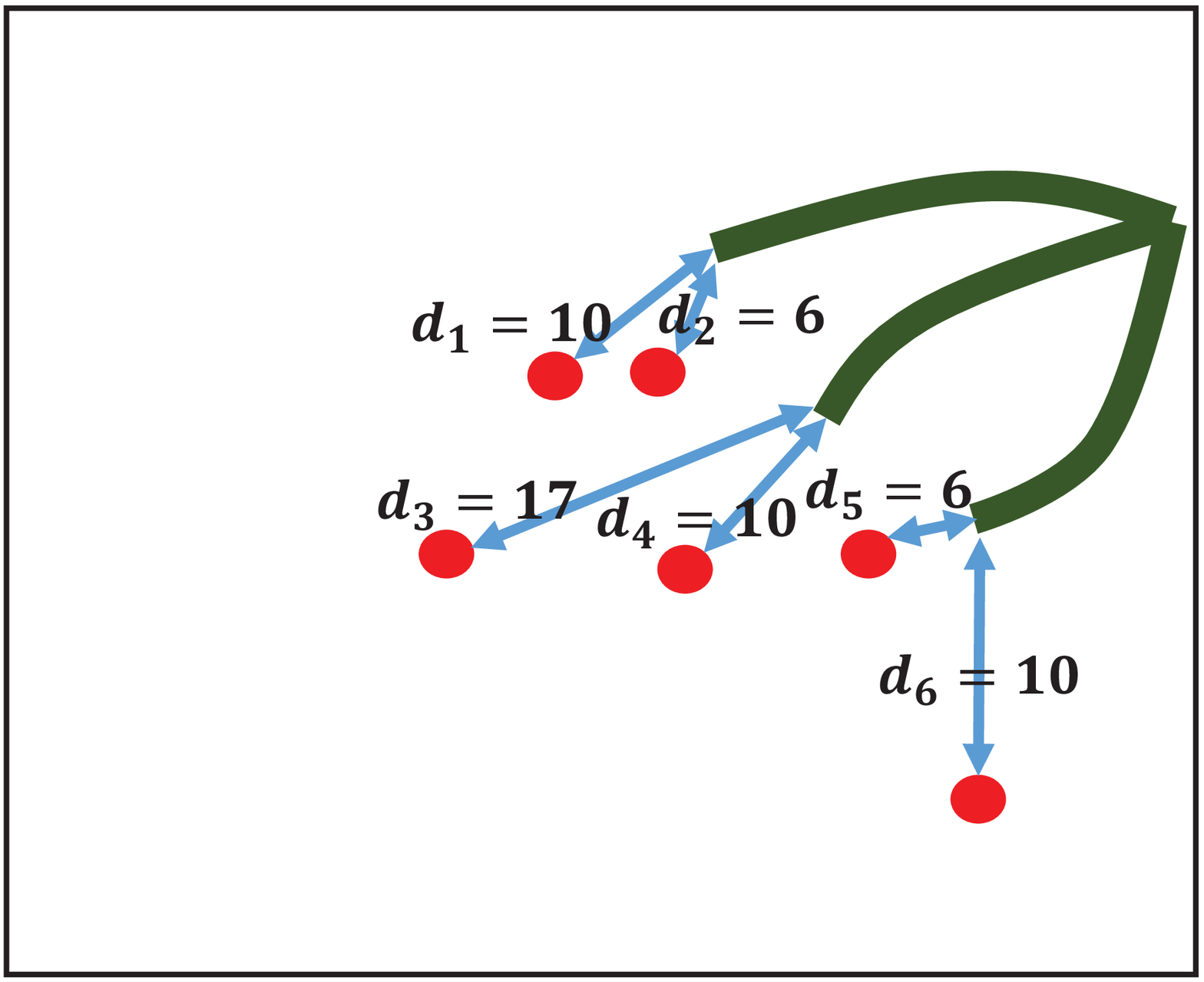}
  }
  \subfigure[]{
  \includegraphics[width=2.25in,height=1.5in]{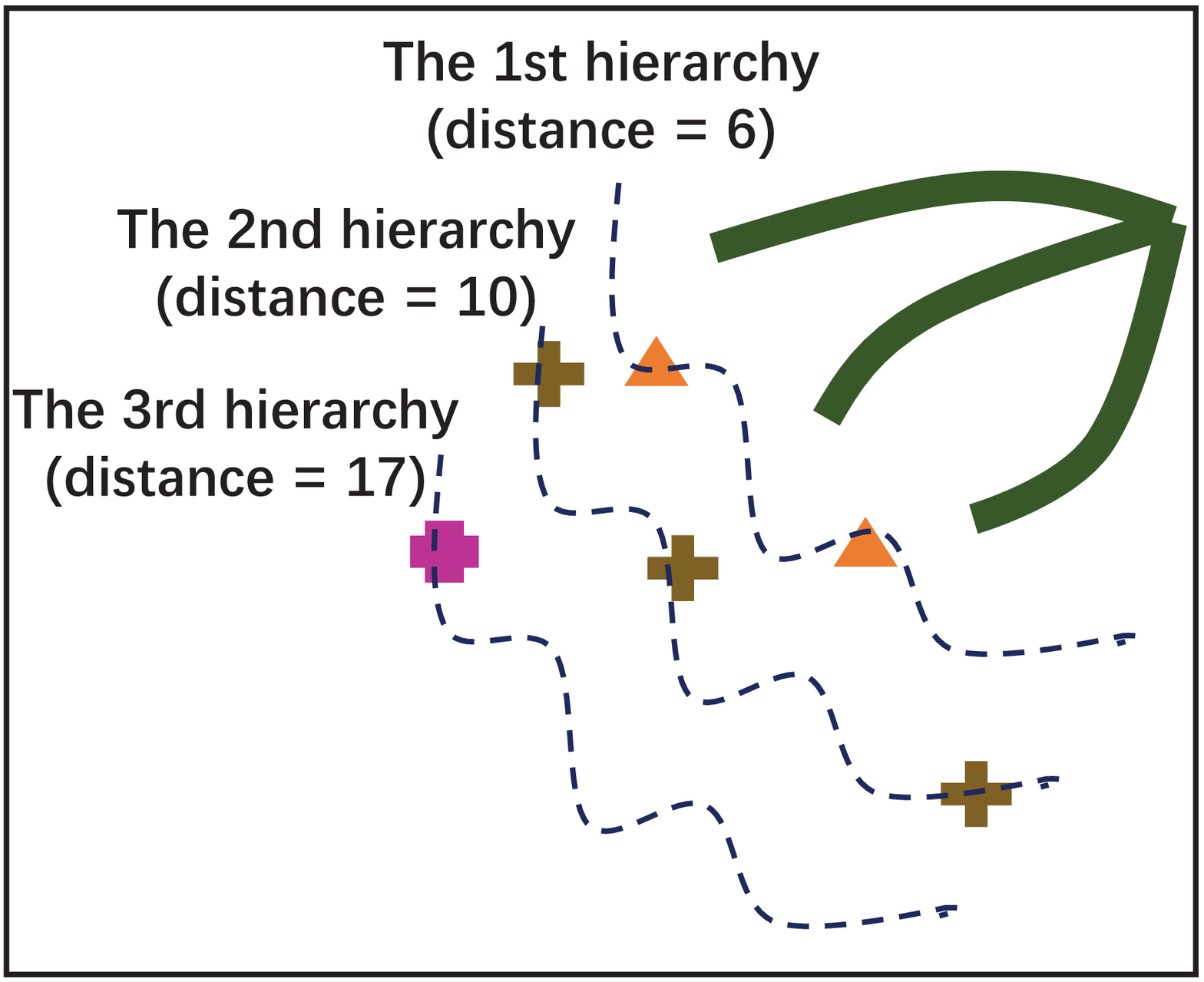}
  }
  \caption{An example to illustrate the process of initialization. (a) An exemplary image (green pixels represent vessel
pixels, red pixels represent unknown pixels). (b) Calculating the closest distance for each unknown pixel ($d_i$ means the closest distance for the $i$th unknown pixel). (c) Stratifying unknown pixels into different hierarchies.  }
\label{fig:initialization}
\end{figure*}
where $t = Otsu(I_{iuw})-\varepsilon$, $\varepsilon$ is set as $0.03$. Then $T$ is divided into three regions according to the $Area$ feature:
\begin{equation}
           T= \left\{
                \begin{array}{rl}
                 T_1 & \text{if $0<Area<a_{1}$} \\
                 T_2 & \text{if $a_{1}\leq Area\leq a_{2}$} \\
                 T_3 & \text{if $a_2<Area$}
                 \end{array} \right.
\end{equation}

In vessel skeleton extraction, the regions in $T_3$ are preserved while the regions in $T_1$ are abandoned. Then the regions in $T_2$ with $Extent> e_2$ \&\& $VRatio\leq r$ are preserved as $T_4$. Finally skeleton extraction \cite{lam1992thinning} is performed on the combined regions of $T_3$ and $T_4$ in order to obtain the skeleton of blood vessels $S$. An example of vessel skeleton extraction is shown in Fig.\ref{fig:vesselSkeleton}.

\begin{figure}
  \centering
  \subfigure[]{
  \includegraphics[width=0.95in]{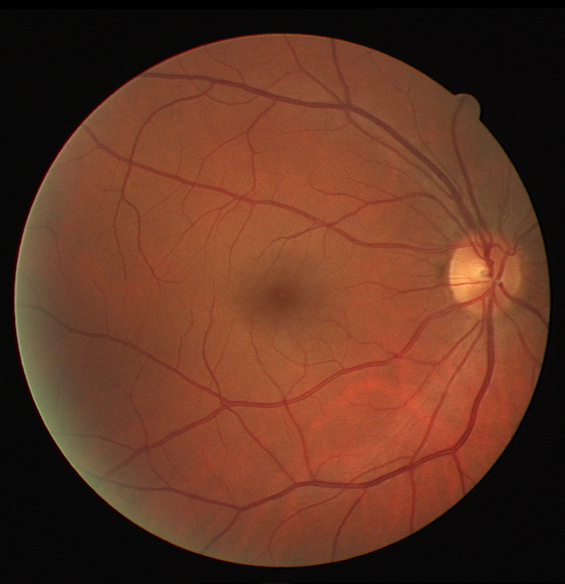}
  }
  \subfigure[]{
  \includegraphics[width=0.95in]{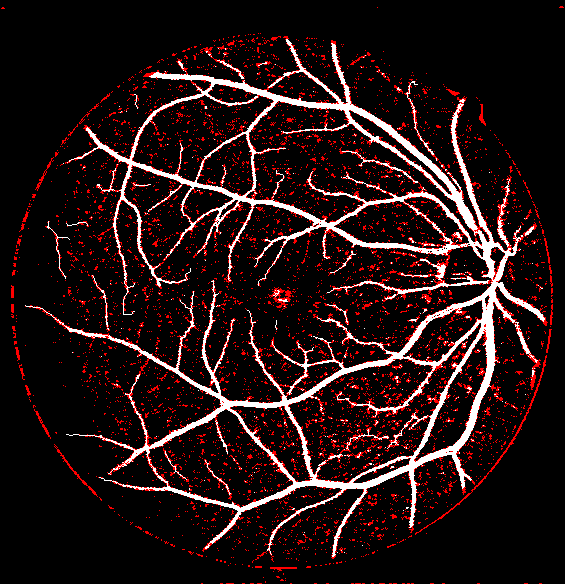}
  }
  \subfigure[]{
  \includegraphics[width=0.95in]{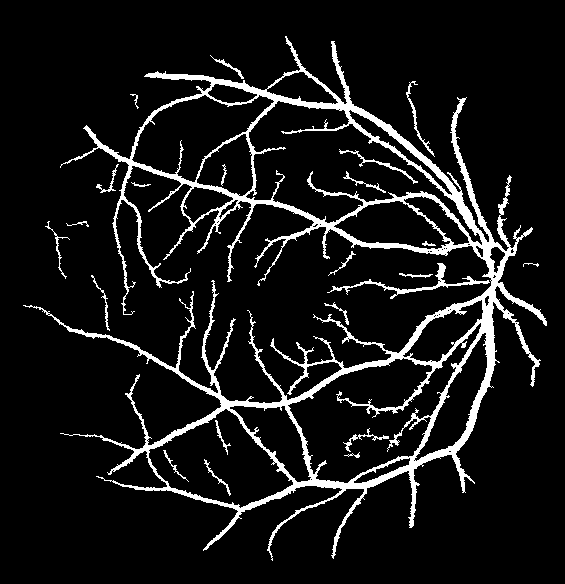}
  }
  \caption{(a) An input image. (b) A trimap generated by the proposed method. The white, black and red regions belong to the foreground, background and unknown regions, respectively. (c) The result achieved by the proposed hierarchical image matting model.}
  \label{processImage}
\end{figure}

After performing image segmentation and vessel skeleton extraction, the trimap of the input fundus image is generated (as shown in Fig.\ref{processImage}(b)), which is composed of the background regions ($B$), unknown regions ($U$) and vessel (or foreground) regions ($V=V_2\cup S$).

\subsection{Hierarchical Image Matting Model}
Hierarchical image matting model is proposed to label the pixels in the unknown regions as vessels or background in an incremental way. Specifically, after stratifying the pixels in unknown regions (called unknown pixels) into $m$ hierarchies by a hierarchical strategy, let $u_i^j$ indicates the $i$th unknown pixel in the $j$th hierarchy, the segmented vessel image $I_v(u_i^j)$ is modeled as follows:
 \begin{equation}
           I_v(z)= \left\{
                \begin{array}{rl}
                 1 & \text{if $corre(u_i^j,V)>corre(u_i^j,B)$} \\
                 0 & else
                 \end{array} \right.
\end{equation}
where $corre$ indicates the correlation function (depicted in Equation \eqref{equ:correlation}). The implementation of the hierarchical image matting model consists of two main steps:
\begin{description} \itemsep -2pt
\item[Step 1]   \emph{Stratifying the unknown pixels:} Stratify pixels in the unknown regions into different hierarchies.
\item[Step 2]   \emph{Hierarchical update:} Assign new labels ($V$ or $B$) to pixels in each hierarchy.
\end{description}
The pseudocode of implementing this model is given in Algorithm 1.

\vspace{1mm}

\scalebox{0.965}{
\begin{tabular}{p{8.5cm}}
\hline
\textbf{Algorithm 1}: Implementing the hierarchical image matting model \\
\hline
  \textbf{Input:} Trimap composed of $B$, $U$, $V$ \\
  \textbf{Output:} The segmented vessel image $I_v$\\
  \textbf{Step 1: Stratifying the unknown pixels:}
  \begin{enumerate}
  \item[a)]For $i = 1, \dots, n_{U}$, set $D(i)=d_i$, where $n_{U}$ is the number of unknown pixels in $U$, $d_i$ is the distance between the $i$th unknown pixel and the closest vessel pixel in $V$, $D$ is the set of $d_i$.
  \item[b)]Sort the unknown pixels in $U$ in an ascending order according to the distances $D$ , cluster the pixels with the same distance into one hierarchy, stratify the pixels into $m$ hierarchies and denote them as an hierarchical order set:
       $H=\{H_1,H_2,\dots,H_m\}$, $H_j=\{u_i^j | i\in{1,2,\dots,n_i}\}$, where $n_i$ is the number of unknown pixels in the $j$th hierarchy $H_j$.
\end{enumerate}

  \textbf{Step 2: Hierarchical Update}\\
  For $j=1,\dots,m$, do \\
   \quad For $i=1,\dots,n_i$, do
  \begin{enumerate}
  \item[a)]Compute the correlations (Defined in Equation \eqref{equ:correlation}) between $u_i^j$ and its neighboring labelled pixels(vessel pixels and background pixels) included in a $9\times9$ grid.
  \item[b)]Choose the labelled pixel with the closest correlation, and assign its label ($V$ or $B$) to $u_i^j$.
  \end{enumerate}
  \quad end for \\
  end for \\
\hline
\end{tabular}}
\vspace{1mm}

\emph{\textbf{Stratifying the unknown pixels:}} In this stage, the unknown pixels are stratified into different hierarchies. For the $i$th unknown pixel in $U$, its distances with all vessel pixels in $V$ are calculated first. Then the closest distance $d_i$ is chosen and assigned to the $i$th unknown pixel. After that, the unknown pixels are stratified into different hierarchies according to the closest distances. The first hierarchy has the lowest value of the closet distance while the last hierarchy has the highest value of the closet distance. The unknown pixels reside in low hierarchy suggests that they are close to blood vessels; The unknown pixels stay in high hierarchy indicates that they are far away from blood vessels. An example to illustrate the process of stratifying the unknown pixels is shown in Fig.\ref{fig:initialization}.

\begin{figure}
  \centering
  \subfigure[]{
	\includegraphics[width=1.62in,height=1.3in]{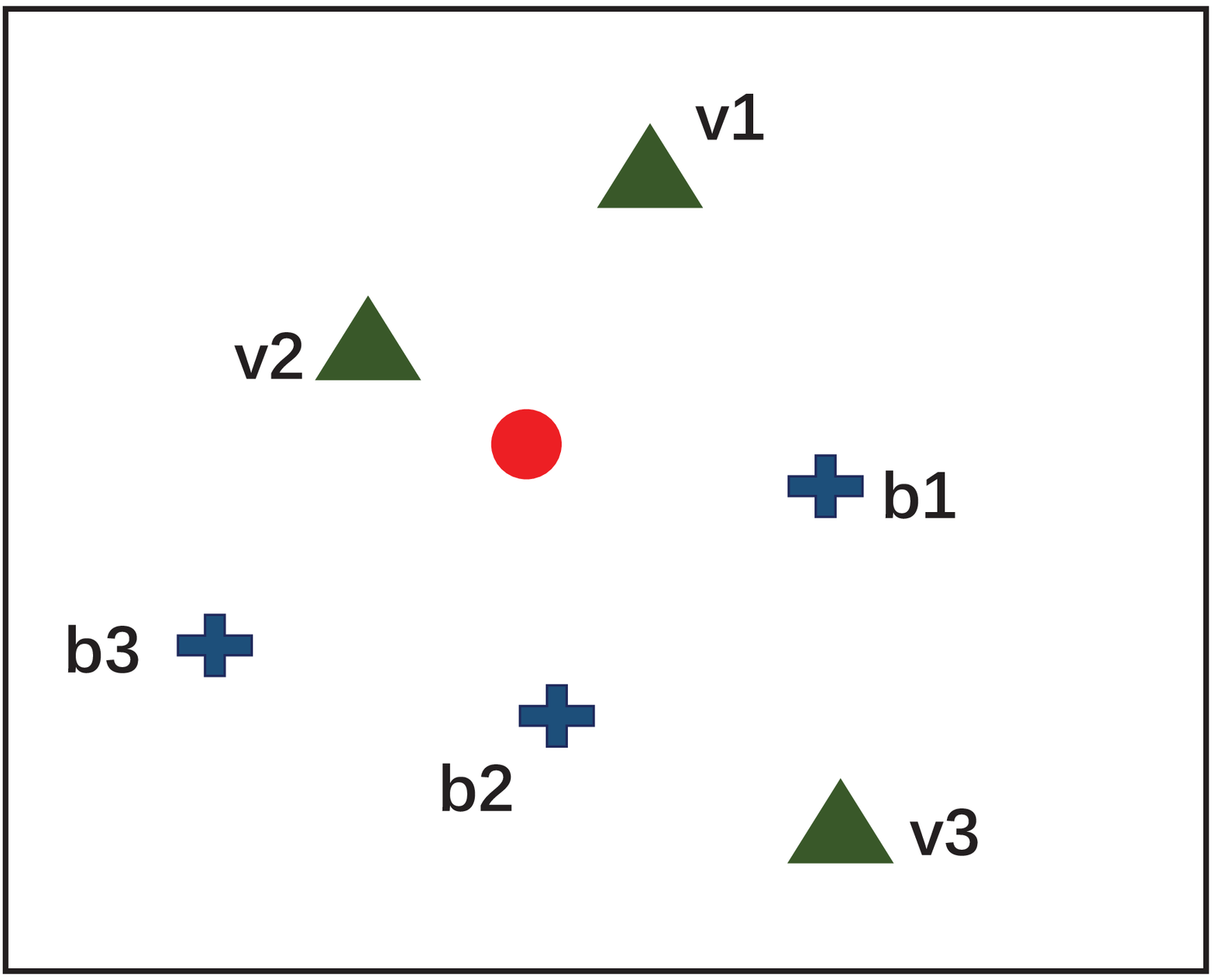}
  }
  \subfigure[]{
	\includegraphics[width=1.62in,height=1.3in]{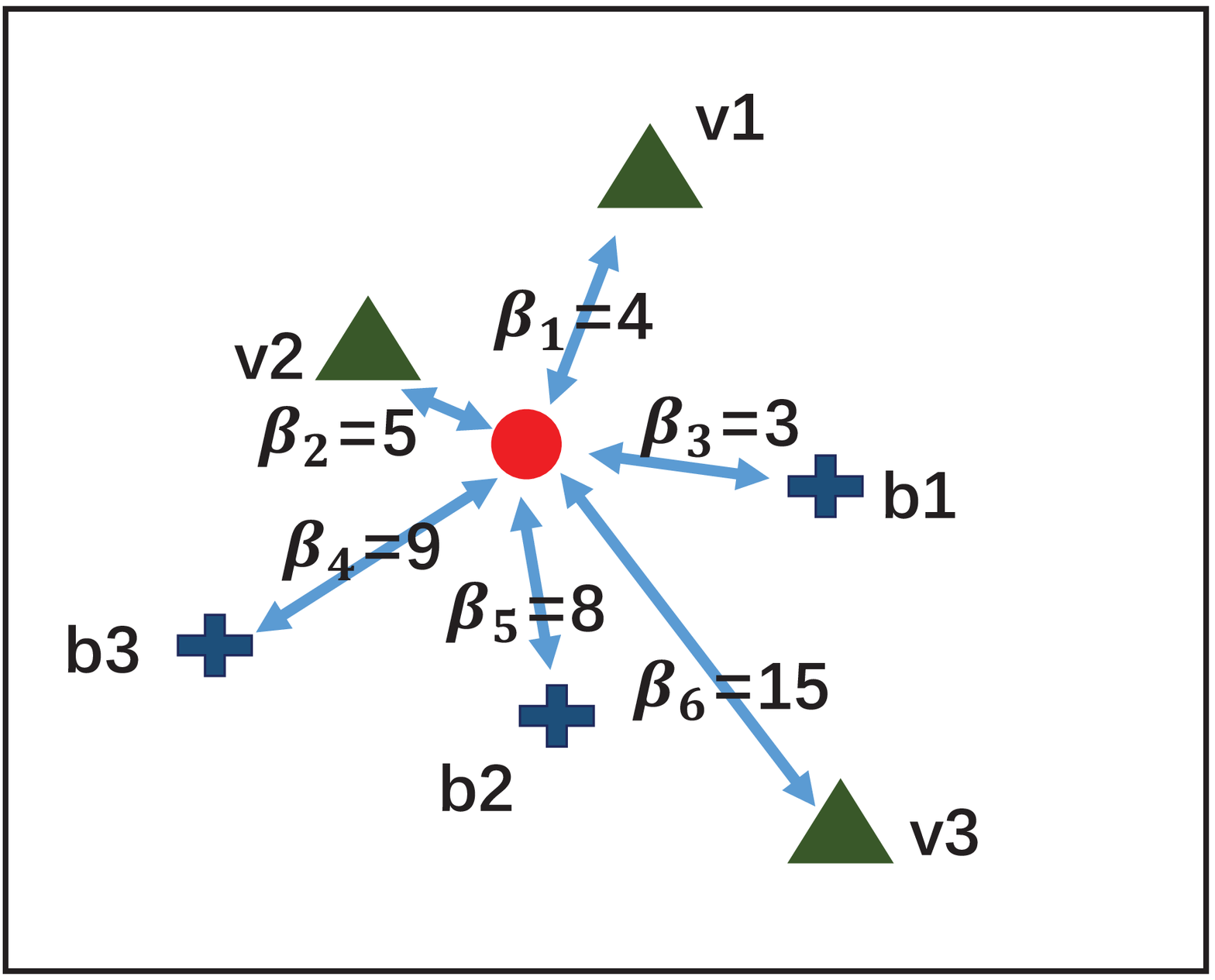}
  }
    \subfigure[]{
	\includegraphics[width=1.62in,height=1.3in]{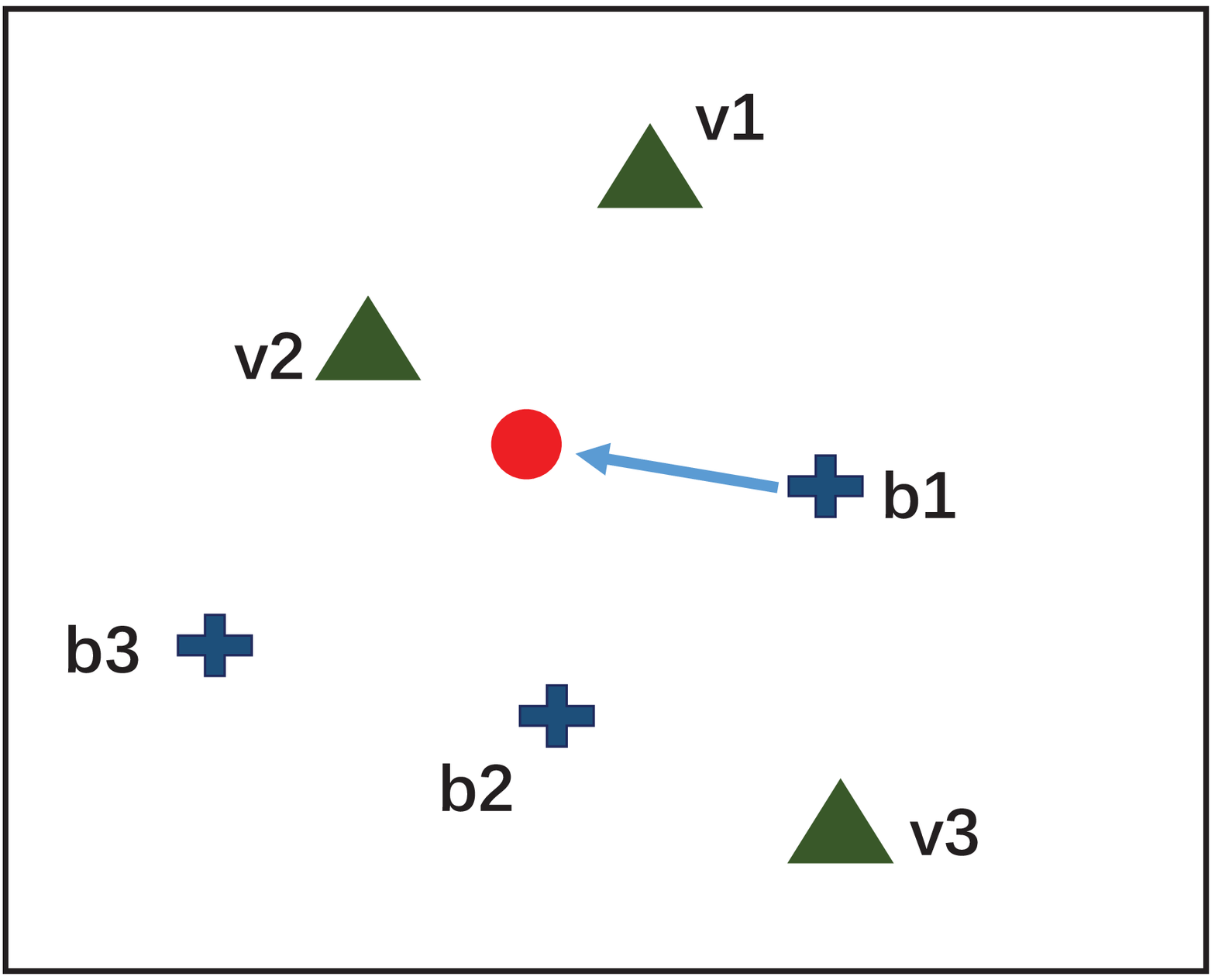}
  }
    \subfigure[]{
	\includegraphics[width=1.62in,height=1.3in]{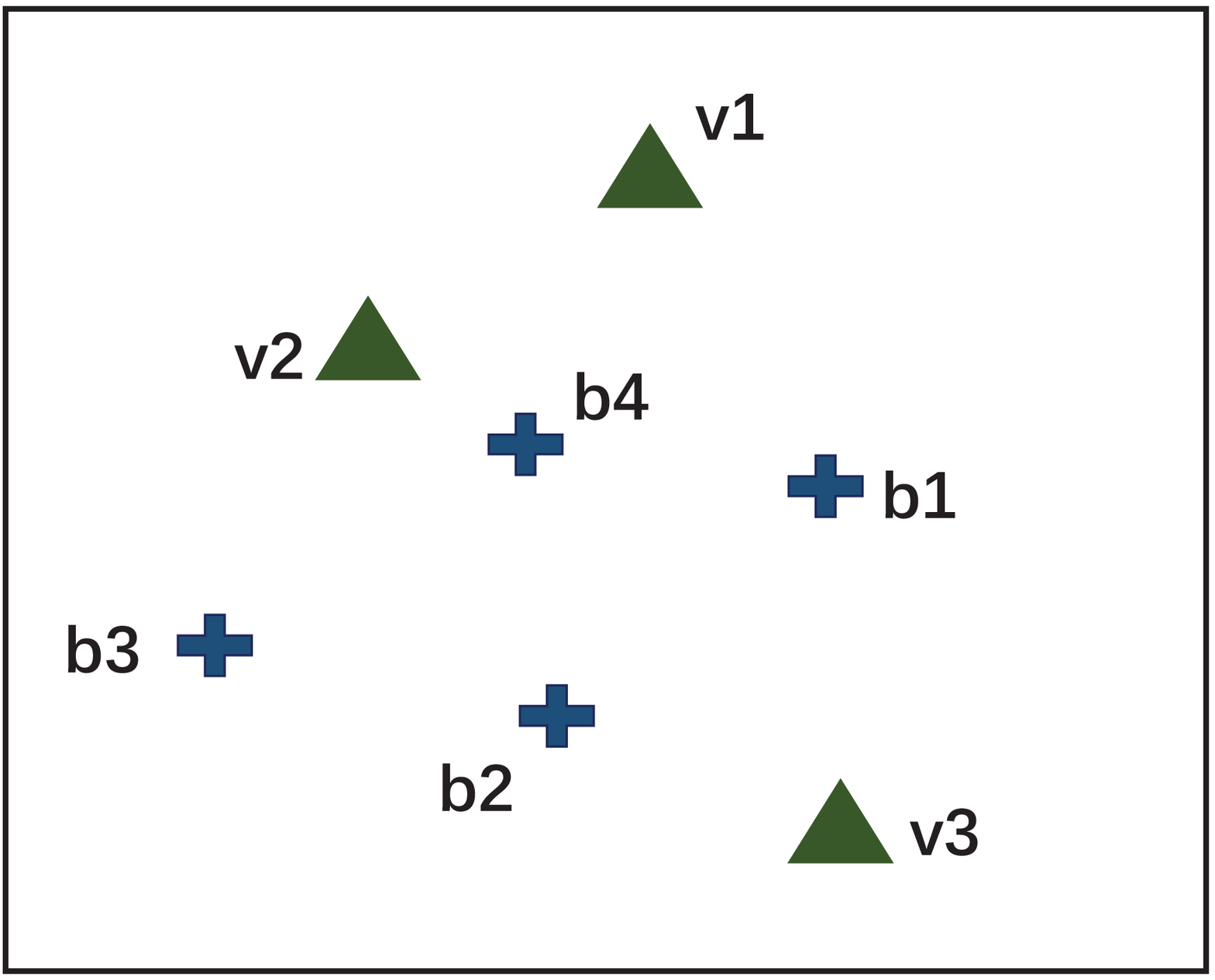}
  }
  \caption{ An example for the illustration of assigning a label ($V$ or $B$) to an unknown pixel. (a) An exemplary image (green triangles represent vessel pixels, blue pluses represent background pixels, red points represent unknown pixels). v$i$ indicates the $i$th vessel pixel, b$i$ indicates the $i$th background pixel. (b) Calculating the correlation functions between a unknown pixel and its neighboring labelled pixels (vessel pixels and background pixels) ($\beta_i$ means the correlation between the unknown pixel and the $i$th labelled pixel). (c) Assigning a label ($V$ or $B$) to the unknown pixel. (d) The resultant image.}
  \label{fig:update1}
\end{figure}

\emph{\textbf{Correlation Function:}} In step 2 of Algorithm 1, given an unknown pixel $u_i^j$ and its neighboring labelled pixel $k_l^j$, a color cost function $\beta_c$ is defined to describe the fitness of $u_i^j$ and $k_l^j$ first:
\begin{equation}\label{equ:color1}
\beta_c(u_i^j,k_l^j) = || c_{u_i^j}-c_{k_l^j} ||
\end{equation}
where $c_{u_i^j}$ and $c_{k_l^j}$ are intensity level of $u_i^j$ and $k_l^j$ in $I_{mr}$. A spatial cost function $\beta_s$ is further defined:
\begin{equation}\label{equ:color}
\beta_s(u_i^j,k_l^j)=\frac{||x_{u_i^j}-x_{k_l^j}||- x_{min}}{x_{max}-x_{min}}
\end{equation}
where $x_{u_i^j}$ and $x_{k_l^j}$ are the spatial coordinates of $u_i^j$ and $k_l^j$. The terms $x_{max}=max_j||x_{u_i^j}-x_{k_l^j}||$ and $x_{min}=min_j||x_{u_i^j}-x_{k_l^j}||$ are the maximum and minimum distance of the unknown pixel $u_i^j$ to the labelled pixel $k_l^j$. The normalization factors $x_{min}$ and $x_{max}$ ensure that $\beta_s$ is independent from the absolute distance.

Our final correlation function $\beta$ is a combination of the color fitness and the spatial distance:
\begin{equation}\label{equ:correlation}
\beta(u_i^j,k_l^j) = \beta_c(u_i^j,k_l^j) + \omega\beta_s(u_i^j,k_l^j)
\end{equation}
where $\omega$ is a weight to trade off the color fitness and spatial distance. $\omega$ is set as $0.5$ in our experiment. Intuitively, a small $\beta$ indicates that the labelled pixel has a close correlation with the unknown pixel.

\emph{\textbf{Hierarchical Update:}} After performing initialization with the hierarchical strategy, in each hierarchy, the correlations between each unknown pixel and its neighboring labelled pixels (vessel pixels and background pixels) included in a $9\times9$ grid are computed. Then the labelled pixel with the closest correlation is chosen, and its label is assigned to the unknown pixel. After all unknown pixels in one hierarchy are updated, they are used for the update of the next hierarchy. The unknown pixels are updated from the first hierarchy to the last hierarchy. An example to illustrate the process of updating unknown pixels in one hierarchy is shown in Fig.\ref{fig:update1}.

\subsection{Postprocessing}
Since some non-vessel regions may still exist in the final segmented vessel image $I_v$, the regions whose $Area< a_2$ \&\& $ Extensibility>e_2$ \&\& $VRatio<r$ in $I_v$ are abandoned to remove these non-vessel regions.

\section{Datasets and Evaluation Metrics}
In this section, three publicly available datasets are introduced. These datasets have been widely used by researchers to test their own vessel segmentation algorithms. Then some commonly used evaluation metrics are presented, which are also utilized in our experiment and to compare the proposed method with other state-of-art methods.

\subsection{Datasets}
The proposed model is evaluated on three standard datasets: DRIVE \cite{staal2004ridge}, STARE \cite{hoover2000locating} and CHASE\_DB1 \cite{fraz2012ensemble}.

\textbf{DRIVE}\footnote{http://www.isi.uu.nl/Research/Databases/DRIVE/} consists of 40 fundus images obtained from a screening program in the Netherlands. These images are captured by a Canon CR5 non-mydriatic 3-CCD camera at $45^\circ$ field of view (FOV), and the size of each image is of $584\times 565$ pixels. The DRIVE dataset is divided into two sets: a training set (DRIVE Training) and a test set (DRIVE Test) each containing 20 fundus images. The training set is annotated by two observers; The test set is annotated by two independent human observers.

\textbf{STARE}\footnote{http://www.ces.clemson.edu/~ahoover/stare/} consists of 20 fundus images. These images are captured by a TopCon TRV-50 fundus camera at $35^\circ$ FOV, and the size of each image is of $605\times 700$ pixels. Ten images contain pathology while the other ten images are normal. The STARE dataset is annotated by two independent observers.

%Only the manual segmentations of the first observer are used to validate the proposed model, a common choice for this dataset (e.g., in \cite{zhao2015automated} and \cite{annunziata2016leveraging} ).

\textbf{CHASE\_DB1}\footnote{https://blogs.kingston.ac.uk/retinal/chasedb1/} consists of 28 fundus images acquired from multiethnic school children. These images are captured by a hand-held Nidek NM-200-D fundus camera at $30^\circ$ FOV, and the size of each image is of $960\times 999$ pixels. The CHASE\_DB1 is annotated by two independent observers.

For the DRIVE, STARE and CHASE\_DB1 datasets, the manual segmentations of the first observer are used in this work, which is a common choice for these datasets \cite{annunziata2016leveraging,Liskowski2016Segmenting,fraz2012blood,zhao2015automated}.

% Table generated by Excel2LaTeX from sheet 'comparison'
\begin{table*}[htbp]
  \centering
  \footnotesize
  \caption{PERFORMANCE OF DIFFERENT SEGMENTATION MODELS ON THE DRIVE AND STARE DATASETS}
    \begin{tabular}{lcccccccccccc}
    \toprule
    \toprule
    Test Datasets & \multicolumn{4}{c}{DRIVE} &       &       & \multicolumn{5}{c}{STARE}             &  \\
    \midrule
    Methods & Acc   & AUC   & Se    & Sp    & Time  &       & Acc   & AUC   & Se    & Sp    & Time  & System \\
    \midrule
    \textbf{Supervised Methods} &       &       &       &       &       &       &       &       &       &       &       &  \\
    Staal \emph{et.al}\cite{staal2004ridge} & 0.944  & -     & -     & -     & 15min &       & 0.952  & -     & -     & -     & 15min & 1.0 GHz, 1-GB RAM \\
    Soares \emph{et.al}\cite{soares2006retinal} & 0.946  & -     & -     & -     & $\sim$3min &       & 0.948  & -     & -     & -     & $\sim$3min & 2.17 GHz, 1-GB RAM \\
    Lupascu \emph{et.al}\cite{lupascu2010fabc} & 0.959  & -     & 0.720  & -     & -     &       & -     & -     & -     & -     & -     & - \\
    Marin \emph{et.al}\cite{marin2011new} & 0.945  & 0.843  & 0.706  & 0.980  &$\sim$90s  &       & 0.952  & 0.838  & 0.694  & 0.982  & $\sim$90s  & 2.13 GHz, 2-GB RAM \\
    Roychowdhury  \emph{et.al}\cite{roychowdhury2015blood} & 0.952  & 0.844  & 0.725  & 0.962  & 3.11s &       & 0.951  & 0.873  & 0.772  & 0.973  & 6.7s  & 2.6 GHz, 2-GB RAM \\
    Liskowski \emph{et.al}\cite{Liskowski2016Segmenting} & 0.954  & 0.881  & 0.781  & 0.981  & -     &       & 0.973  & 0.921  & 0.855  & 0.986  &  -     & NVIDIA GTX Tian GPU \\
    \midrule
    \textbf{Unsupervised Methods} &       &       &       &       &       &       &       &       &       &       &       &  \\
    Hoover  \emph{et.al}\cite{hoover2000locating} & -     & -     & -     & -     & -     &       & 0.928  & 0.730  & 0.650  & 0.810  & 5min  & Sun SPARCstation 20 \\
    Mendonca  \emph{et.al}\cite{mendonca2006segmentation} & 0.945  & 0.855  & 0.734  & 0.976  & 2.5min &       & 0.944  & 0.836  & 0.699  & 0.973  & 3min  & 3.2 GHz, 980-MB RAM \\
    Lam  \emph{et.al}\cite{lam2008novel} & -     & -     & -     & -     & -     &       & 0.947  & -     & -     & -     & 8min  & 1.83 GHz, 2-GB RAM \\
    Al-Diri  \emph{et.al}\cite{al2009active} & -     & -     & 0.728  & 0.955  & 11min &       & -     & -     & 0.752  & 0.968  & -     & 1.2 GHz \\
    Lam and Yan \emph{et.al}\cite{lam2010general} & 0.947  & -     & -     & -     & 13min &       & 0.957  & -     & -     & -     & 13min & 1.83 GHz, 2-GB RAM \\
    Perez  \emph{et.al}\cite{palomera2010parallel} & 0.925  & 0.806  & 0.644  & 0.967  & $\sim$2min &       & 0.926  & 0.857  & 0.769  & 0.944  & $\sim$2min & Parallel Cluster \\
    Miri  \emph{et.al}\cite{miri2011retinal} & 0.943  & 0.846  & 0.715  & 0.976  & $\sim$50s  &       & -     & -     & -     & -     & -     & 3 GHz, 1-GB RAM \\
    Budai  \emph{et.al}\cite{budai2013robust} & 0.957  & 0.816  & 0.644  & 0.987  & -     &       & 0.938  & 0.781  & 0.580  & 0.982  & -     & 2.3 GHz, 4-GB RAM \\
    Nguyen  \emph{et.al}\cite{nguyen2013effective} & 0.941  & -     & -     & -     & 2.5s  &       & 0.932  & -     & -     & -     & 2.5s  & 2.4 GHz, 2-GB RAM \\
      Yitian  \emph{et.al}\cite{zhao2015automated} & 0.954  & 0.862  & 0.742  & 0.982  & -     &       & 0.956  & 0.874  & 0.780  & 0.978  & -     & 3.1 GHz, 8-GB RAM \\

            Annunziata  \emph{et.al}\cite{annunziata2016leveraging} & -  & -  & -  & -  & -     &       & 0.956  & 0.849  & 0.713  & 0.984  & $<$25s     & 1.9 GHz, 6-GB RAM \\

    Orlando  \emph{et.al}\cite{orlando2017discriminatively} & -  & 0.879  & 0.790  & 0.968  & -     &       & -  &0.871   &0.768   &0.974   & -     & 2.9 GHz, 64-GB RAM \\
      \textbf{Proposed} & \textbf{0.960} & \textbf{0.858}  & \textbf{0.736} & \textbf{0.981}  & \textbf{10.72s} &       & \textbf{0.957} & \textbf{0.880} & \textbf{0.791}  & \textbf{0.970} & \textbf{15.74s} & \textbf{2.5 GHz, 4-GB RAM} \\
    \bottomrule
    \bottomrule
    \end{tabular}%
  \label{tab:com}%
\end{table*}%

\subsection{Evaluation Metrics}
In the process of retinal vessel segmentation, each pixel is classified as vessels or background, thereby resulting in four events: two correct (true) classifications and two incorrect (false) classifications, which are defined in Table \ref{tab:event}.
\begin{table}[htbp]
  \centering
  \caption{FOUR EVENTS OF VESSEL CLASSIFICATION}
    \begin{tabular}{|l|l|l|}
    \hline
          & Vessel present & vessel absent \\
    \hline
    vessel detected & True Positive (TP) & False Positive (FP) \\
    \hline
    vessel not detected & False Negative (FN) & True Negative(TN) \\
    \hline
    \end{tabular}%
  \label{tab:event}%
\end{table}%

In order to evaluate the performance of the vessel segmentation algorithms, three commonly used metrics are applied.
\begin{equation*}
\begin{split}
       Sensitivity &= \frac{TP}{TP+FN}   \\
       Specificity &= \frac{TN}{TN+FP}   \\
       Accuracy &= \frac{TP+TN}{TP+TN+FP+FN}
\end{split}
\end{equation*}
Sensitivity ($Se$) reflects the algorithm's ability of detecting vessel pixels while Specificity ($Sp$) is a measure of the algorithm's effectiveness in identifying background pixels. Accuracy ($Acc$) is a global measure of classification performance combing both $Se$ and $Sp$. The performance of the vessel segmentation method is also measured by the area under a receiver operating characteristic ($ROC$) curve ($AUC$). The conventional $AUC$ is calculated from a number of operating points, and normally used to evaluate the balanced data classification problem. However, in practice the researchers need to select an operating point to compare their method with other methods. Also blood vessel segmentation is an unbalanced data classification problem, in which there are much fewer vessel pixels than the background pixels. In order to evaluate the performance of blood vessel segmentation properly, $AUC=(Se+Sp)/2$ \cite{hong2007kernel,zhao2015automated} is applied to indicate the overall performance of blood vessel segmentation, which is suitable to describe the overall performance of imbalanced data classification problem and specifically for the case when only one operating point is used. The segmentation time required per image in seconds for implementing the proposed segmentation algorithm in MATLAB on a Laptop with Intel Core i7 processor, 2.5GHz and 8GB RAM is also recorded.

% Table generated by Excel2LaTeX from sheet 'Vessel Segmentation Performance'
\begin{table*}[htbp]
  \centering
  \footnotesize
  \caption{THE SEGMENTATION PERFORMANCE OF THE PROPOSED MODEL ON THREE TEST DATASETS}
    \begin{tabular}{ccccccc}
    \toprule
    Dataset & Method & Acc   & AUC   & Se    & Sp    & Time(s) \\
    \midrule
    \multirow{3}[2]{*}{DRIVE } & Trimap (Treating the unknown regions as background regions) & 0.959  & 0.833  & 0.679  & 0.986  & 5.841  \\
          & Model without vessel skeleton extraction & 0.960  & 0.837  & 0.688  & 0.986  & 11.959  \\
          & Proposed & 0.960  & 0.859  & 0.736  & 0.981  & 10.720  \\
    \midrule
    \multirow{3}[2]{*}{STARE} & Trimap (Treating the unknown regions as background regions) & 0.958  & 0.853  & 0.728  & 0.977  & 7.741  \\
          & Model without vessel skeleton extraction & 0.959  & 0.862  & 0.748  & 0.976  & 16.563  \\
          & Proposed & 0.957  & 0.881  & 0.791  & 0.970  & 15.740  \\
    \midrule
    \multirow{3}[2]{*}{CHASE\_DB1} & Trimap (Treating the unknown regions as background regions) & 0.948  & 0.771 & 0.565  & 0.977  & 21.088  \\
          & Model without vessel skeleton extraction & 0.954  & 0.789  & 0.597  & 0.981  & 60.847  \\
          & Proposed & 0.951  & 0.815 & 0.657  & 0.973  & 50.710  \\
    \bottomrule
    \end{tabular}%
  \label{tab:segmentationPerformance}%
\end{table*}%

\begin{figure*}
\begin{tabular}{cc}
\begin{minipage}[t]{0.25\linewidth}
\includegraphics[width=1.5in]{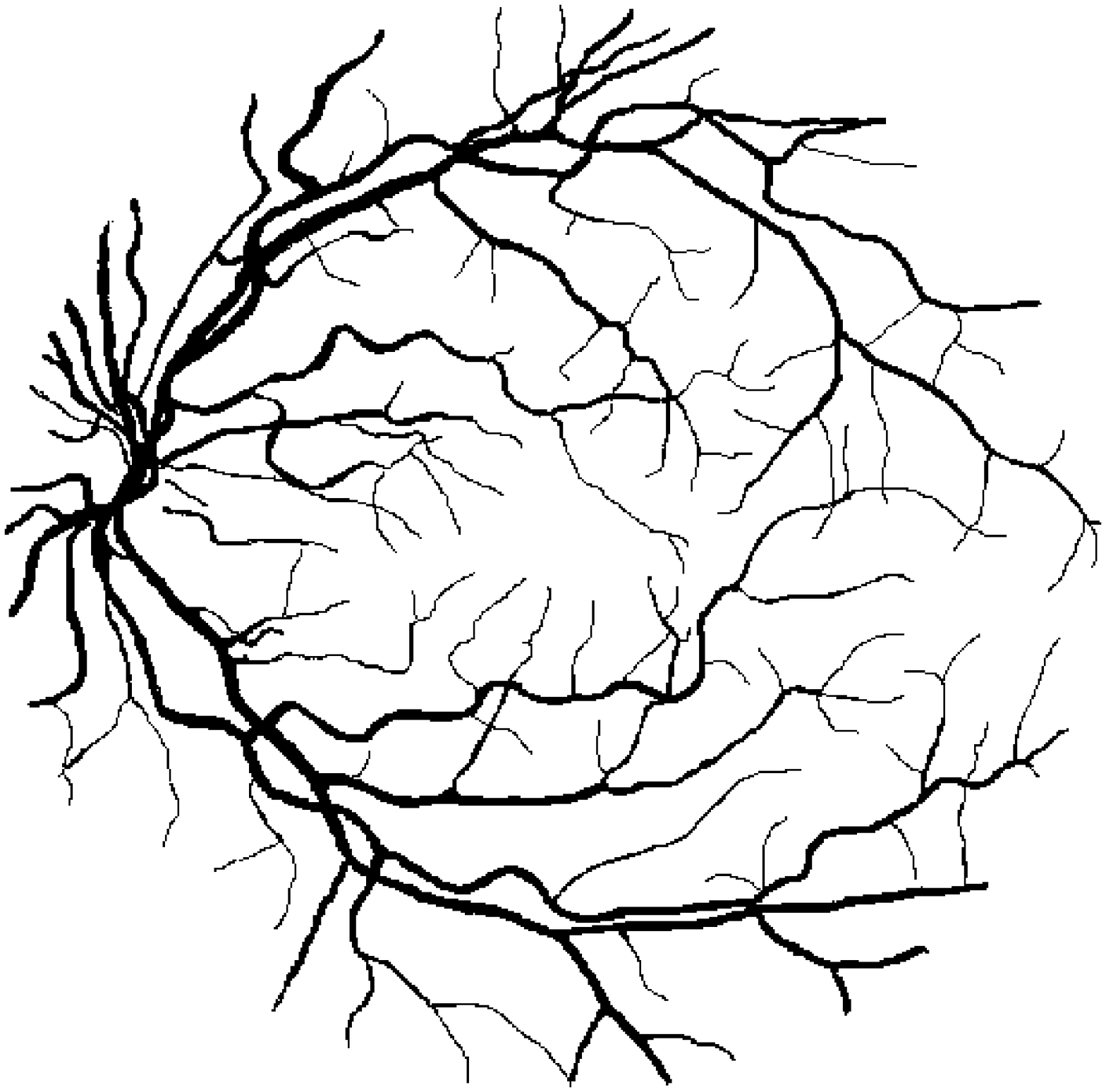}
\end{minipage}
\begin{minipage}[t]{0.25\linewidth}
\includegraphics[width=1.5in]{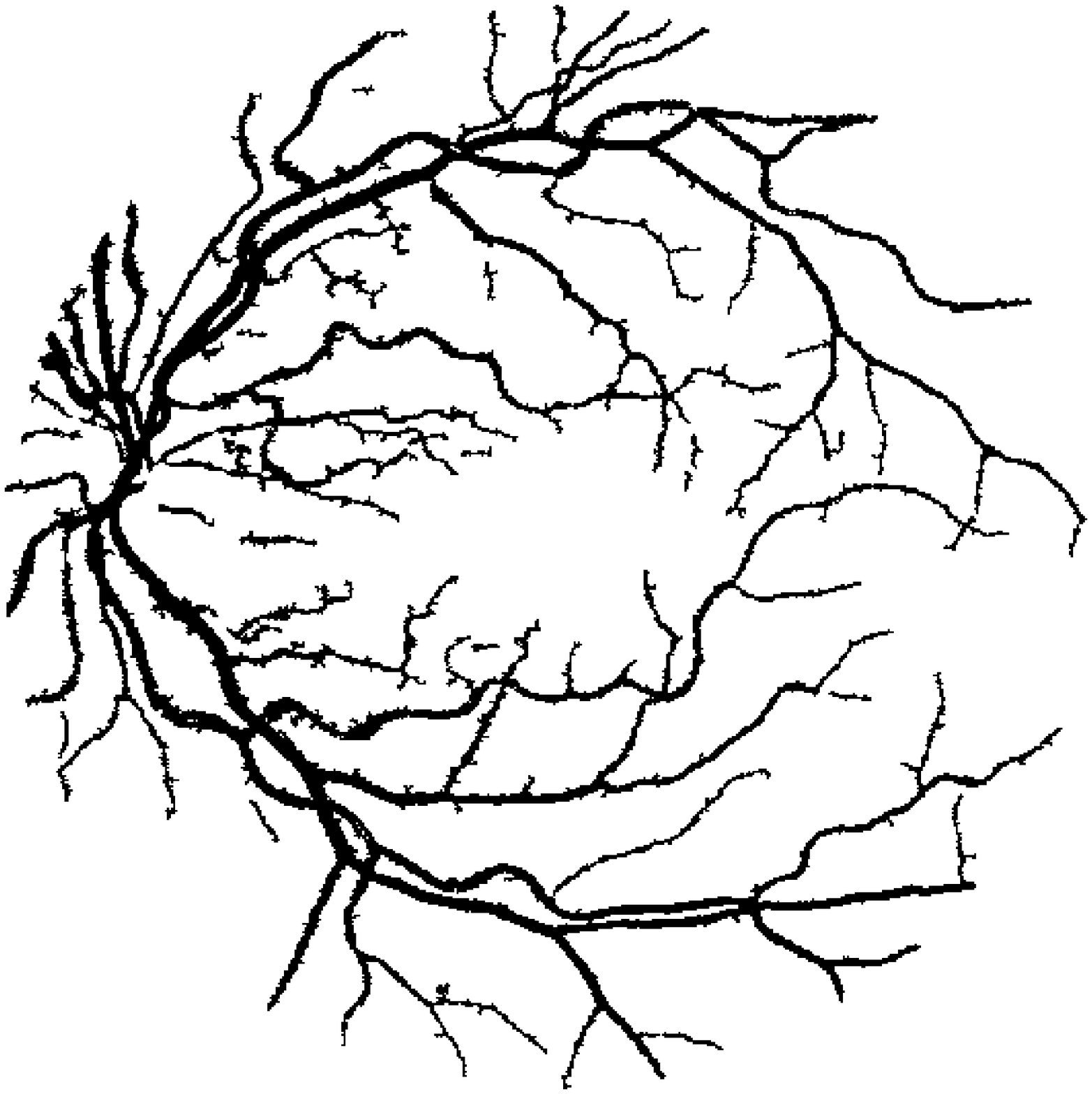}
\end{minipage}
\begin{minipage}[t]{0.25\linewidth}
\includegraphics[width=1.5in]{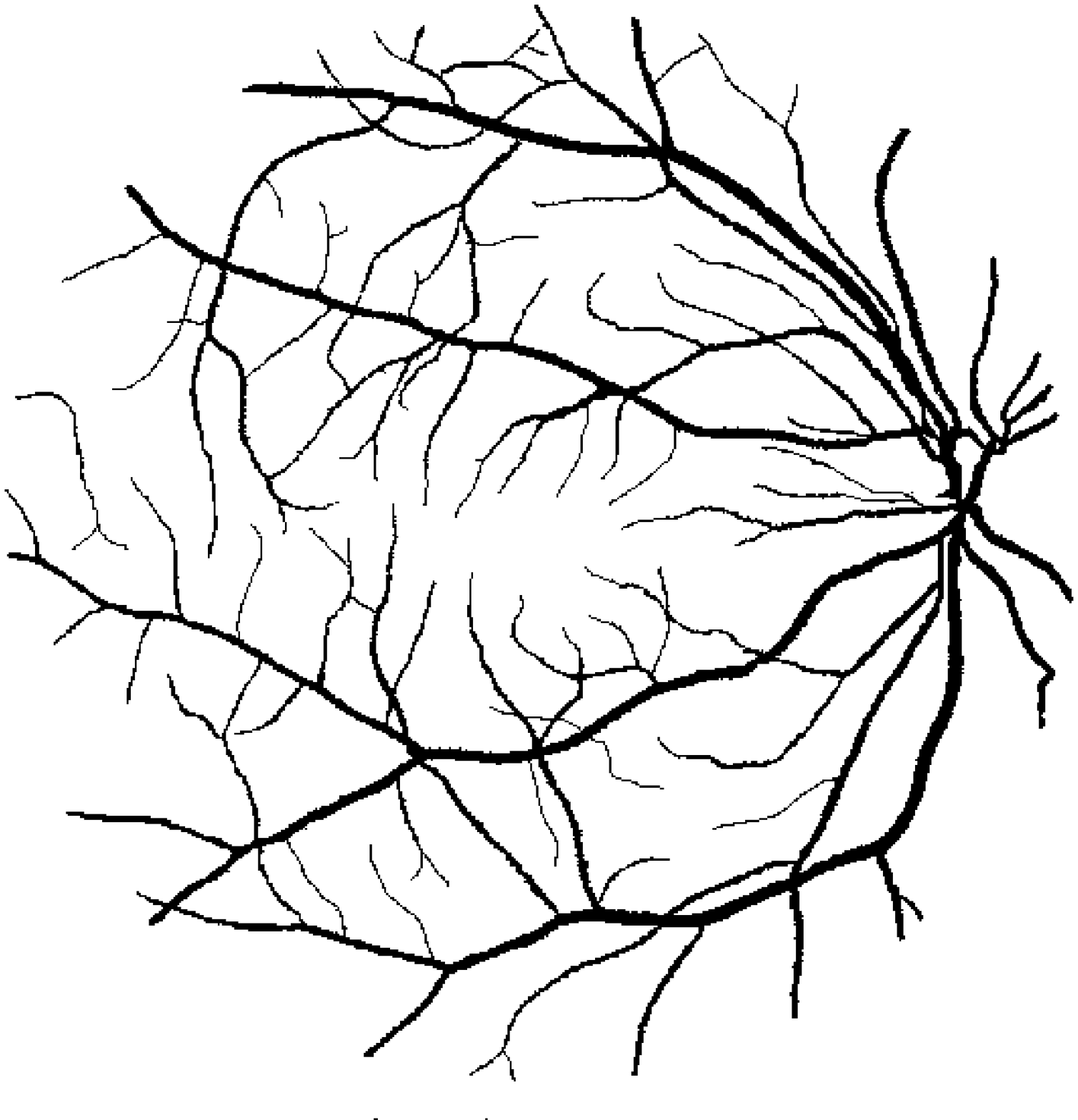}
\end{minipage}
\begin{minipage}[t]{0.25\linewidth}
\includegraphics[width=1.5in]{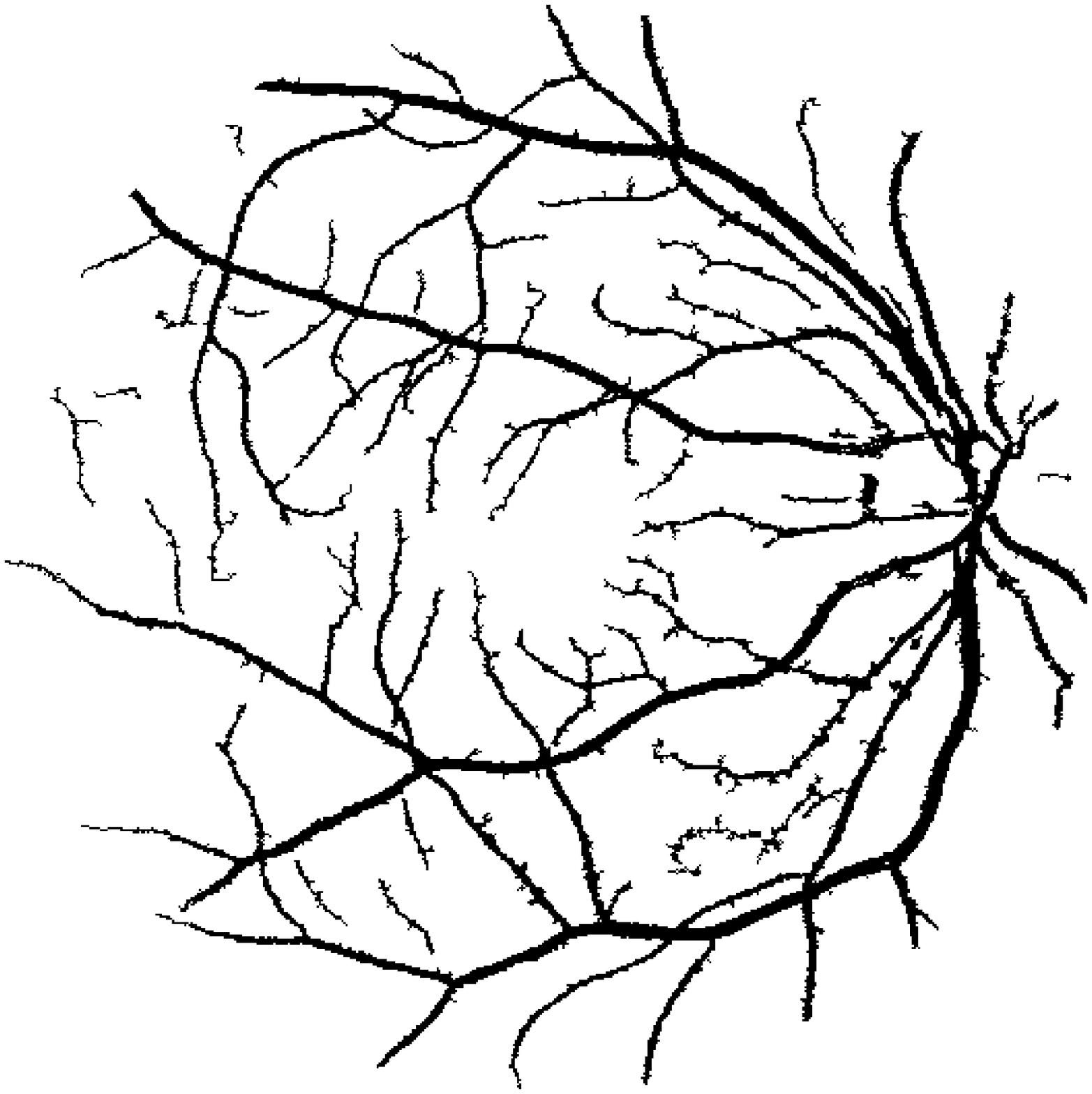}
\end{minipage}
\end{tabular}
\centering{(a) \ \ \ \ \ \ \ \ \ \ \ \ \ \ \ \ \ \ \ \ \ \ \ \ \ \ \ \ \ \ \ \ \ \ \ \ \ \ \ \ \ \ \ \ \ \ \ \ \ \ \ \ \ \ \ \ \ \ \ \ \ \ \ \ \ \ \ \ \ \ \ \  (b)}

\begin{tabular}{cc}
\begin{minipage}[t]{0.25\linewidth}
\includegraphics[width=1.5in]{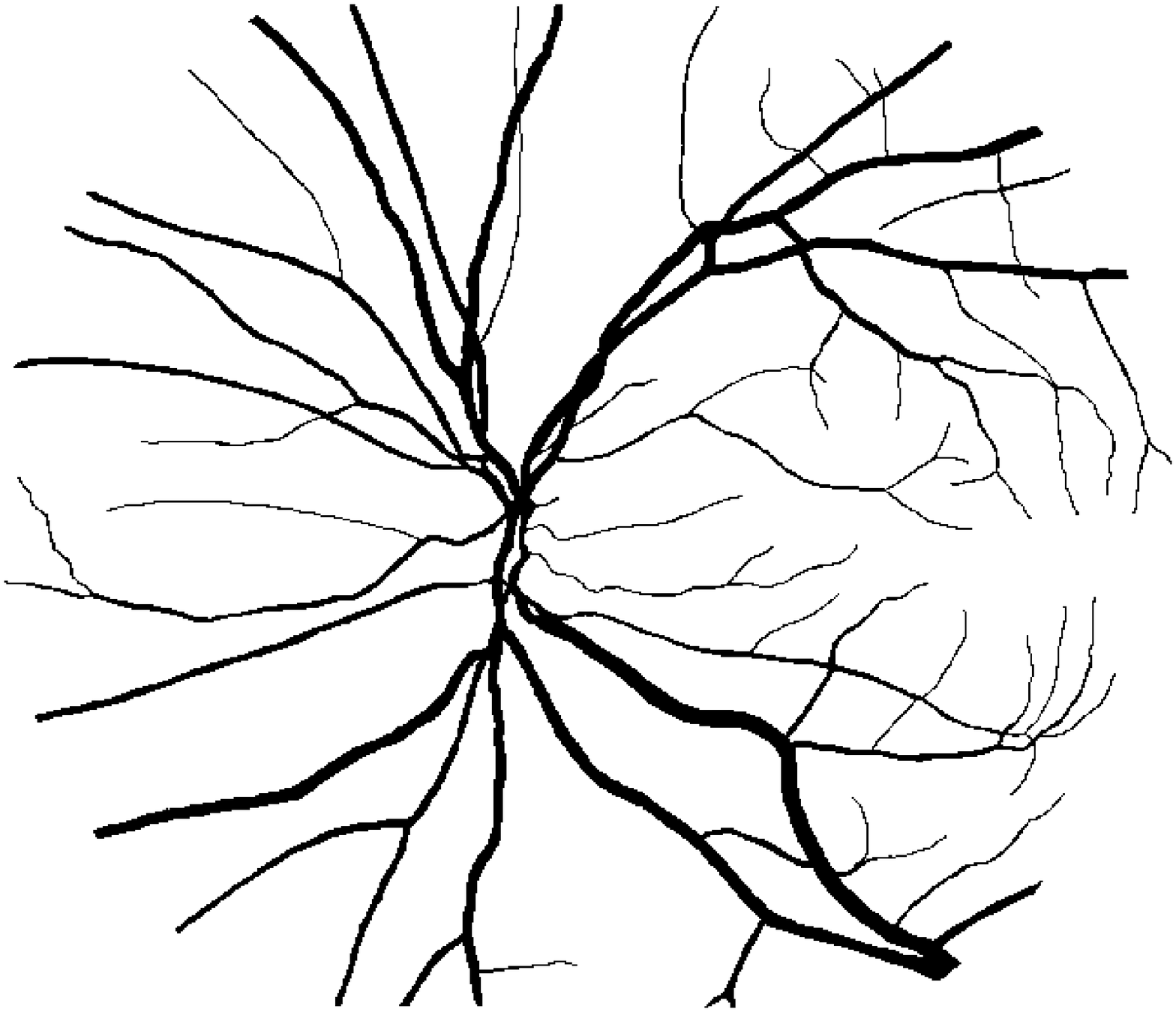}
\end{minipage}
\begin{minipage}[t]{0.25\linewidth}
\includegraphics[width=1.5in]{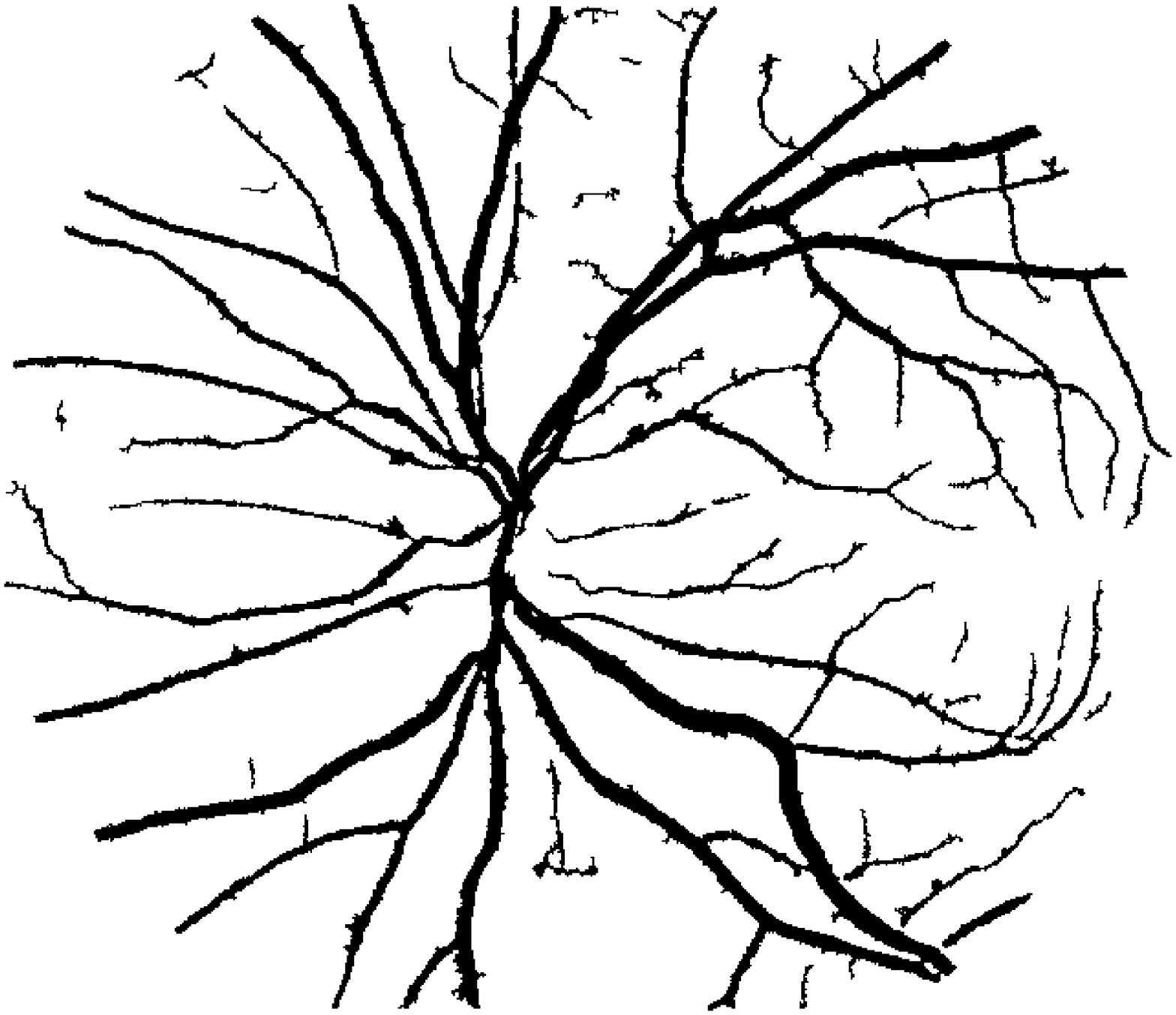}
\end{minipage}
\begin{minipage}[t]{0.25\linewidth}
\includegraphics[width=1.5in]{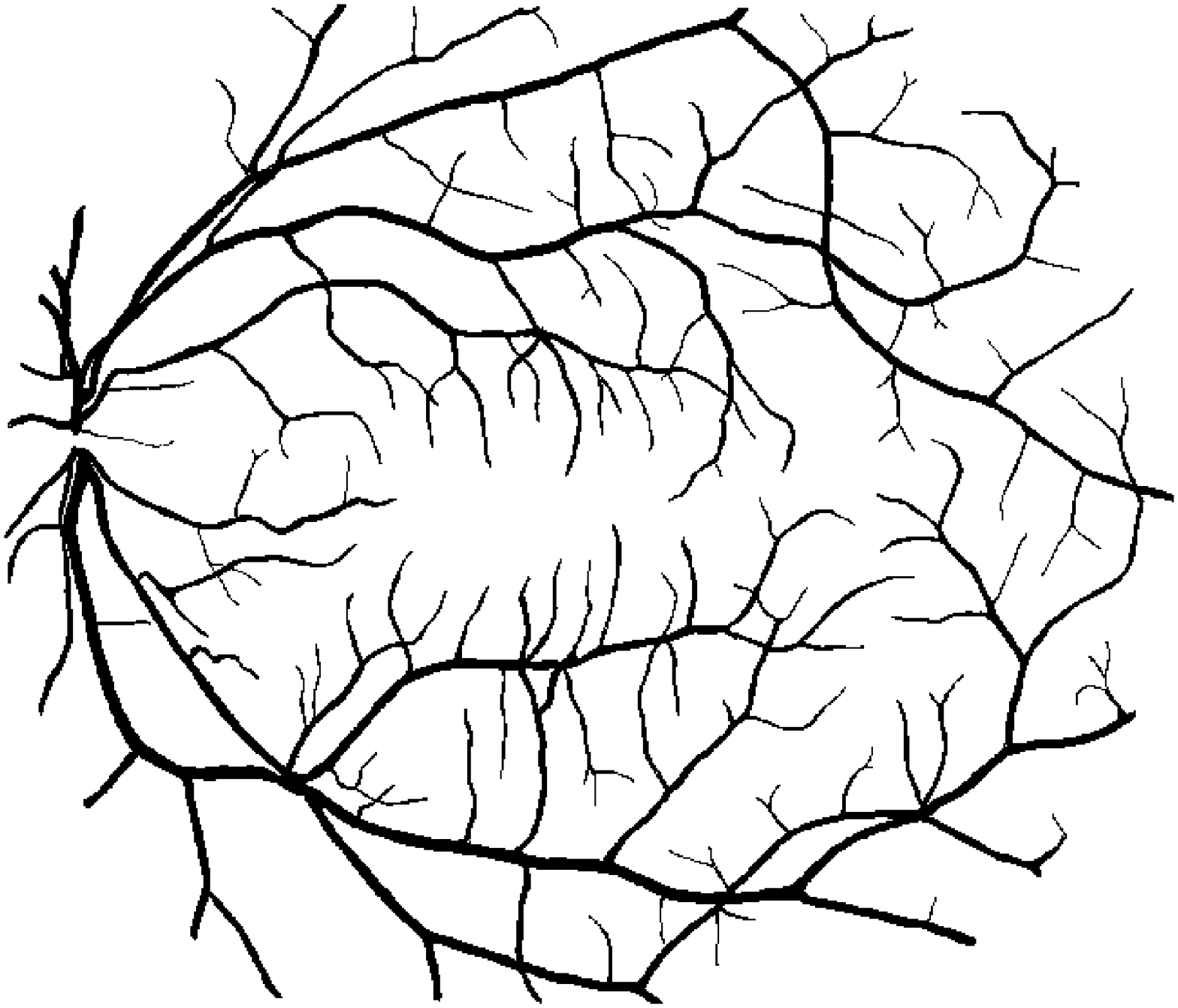}
\end{minipage}
\begin{minipage}[t]{0.25\linewidth}
\includegraphics[width=1.5in]{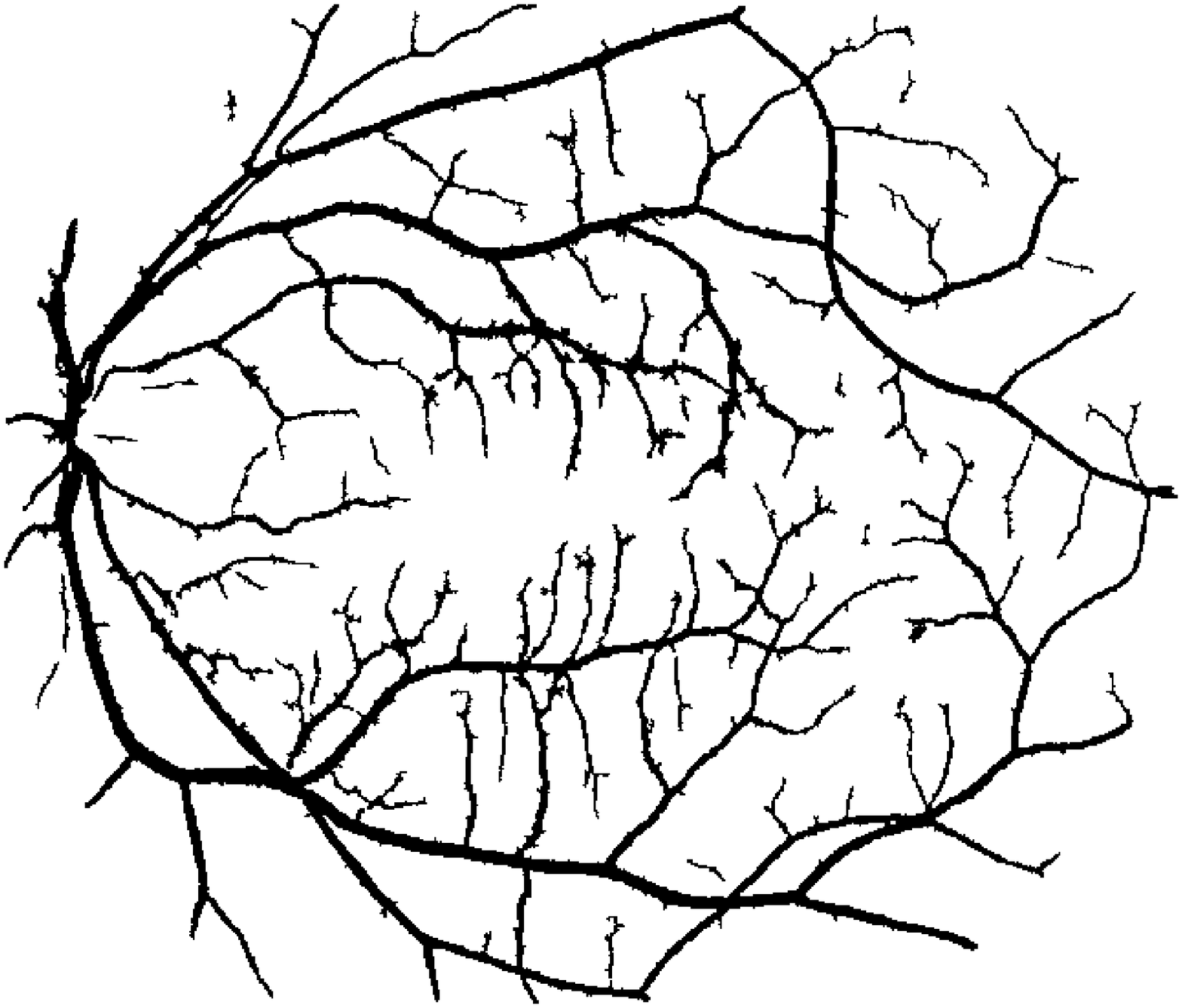}
\end{minipage}
\end{tabular}
\centering{(c) \ \ \ \ \ \ \ \ \ \ \ \ \ \ \ \ \ \ \ \ \ \ \ \ \ \ \ \ \ \ \ \ \ \ \ \ \ \ \ \ \ \ \ \ \ \ \ \ \ \ \ \ \ \ \ \ \ \ \ \ \ \ \ \ \ \ \ \ \ \ \ \  (d)}

  \caption{Ground truth (left) and segmentation result (right): (a) and (b) are the images from DRIVE dataset, (c) and (d) are the images from the STARE dataset}
  \label{fig:finalImage}

\end{figure*}

\section{Experiments and Results}
In this section, four sets of experiments are performed with the purpose of evaluating the proposed hierarchical image matting model. In the first experiment, the segmentation performance of the proposed model was compared with other state-of-art methods. In the second experiment, the vessel segmentation performance of the proposed model was analyzed. In the third experiment, the proposed hierarchical image matting model was compared with image matting models. In the last experiment, the sensitivity analysis of threshold values of region features used in the work was given.

\subsection{Comparison with other methods}
In this section, the proposed model is compared with other state-of-art methods on two most popular public datasets: DRIVE and STARE. The CHASE\_DB1 dataset is not used here since it is relatively
new and has relatively few results in the literature. The segmentation performance and computational complexity of the proposed model in comparison with other state-of-art methods on the DRIVE and STARE datasets are given in Table \ref{tab:com}. It can be observed that for the DRIVE dataset, the accuracy of the proposed model is the highest among all existing methods with $Acc=0.960$, $Se=0.736$ and $Sp=0.981$. On the STARE dataset, the accuracy and $AUC$ of the proposed model are the highest among unsupervised methods with $Acc=0.957$, $AUC=0.880$. Although one supervised method \cite{Liskowski2016Segmenting} has the best performance on STARE dataset, the method is computationally more complex due to the use of deep neural networks, which may need retraining for new datasets. In addition, the proposed model has a low computational complexity compared with other segmentation methods.

\subsection{Vessel Segmentation Performance}
The segmentation performance of the proposed model on three public available datasets is given in Table \ref{tab:segmentationPerformance}. It can be observed that the proposed model can achieve more than $95\%$ segmentation accuracy on the DRIVE, STARE and CHASE\_DB1 datasets, with the highest accuracy score $Acc=96.0\%$ achieved in the DRIVE dataset. Some exemplary segmentation results are shown in Fig.\ref{fig:finalImage}. When treating the unknown regions as background regions, \emph{AUC=0.833} of trimap is $2.6\%$ lower than the proposed model while \emph{Acc} of trimap is similar to the proposed model. In addition, $Se=0.679$ of trimap is $5.7\%$ lower than the proposed model. These observations show that trimap can already have good segmentation performance, which indicates that the selection of region features is very effective in segmenting blood vessels. From Table \ref{tab:segmentationPerformance}, it can be observed that the model with vessel skeleton extraction can achieve more than $5\%$ increase of $Sensitivity$ and $2\%$ increase of $AUC$ compared with the model without vessel skeleton extraction while $Acc$ of the model with vessel skeleton extraction is similar to the model without vessel skeleton extraction, which demonstrates the effectiveness of vessel skeleton extraction.

\subsection{Comparison with image matting models}
The effectiveness of the proposed model in blood vessel segmentation has been validated through previous experiments. In order to further verify the effectiveness of our model, the proposed model is compared with four other image matting models: Anat Model \cite{levin2008closed}, Zheng Model \cite{zheng2009learning}, Shahrian Model \cite{shahrian2012weighted} and Improving Model \cite{shahrian2013improving}. The segmemtation results of
these models on the DRIVE, STARE, and CHASE\_DB1 datasets are given in Table \ref{tab:comparedModel}. It can be observed that the proposed model outperforms other image matting models in terms of $Acc$ in the DRIVE, STARE and CHASE\_DB1 datasets. Also the proposed model achieves the highest scores of $Sp$ in three datasets.

% Table generated by Excel2LaTeX from sheet 'imageMattingModel'
\begin{table}[htbp]
  \centering
  \scriptsize
  \caption{SEGMENTATION PERFORMANCE OF FOUR DIFFERENT IMAGE MATTING MODELS AND THE PROPOSED MODEL}
    \begin{tabular}{ccccccc}
    \toprule
    Dataset & Model  & Acc   & AUC   & Se    & Sp    & Time(s) \\
    \midrule
    \multirow{5}[2]{*}{DRIVE } & Anat Model  & 0.958  & 0.862  & 0.746  & 0.978  & 10.036  \\
          & Zheng Model  & 0.958  & 0.862  & 0.746  & 0.978  & 11.837  \\
          & Shahrian Model  & 0.921  & 0.903  & 0.881  & 0.925  & 120.552  \\
          & Improving Model & 0.958  & 0.862  & 0.746  & 0.978  & 360.000  \\
          & Proposed & 0.960  & 0.859  & 0.736  & 0.981  & 10.720  \\
    \midrule
    \multirow{5}[2]{*}{STARE} & Anat Model  & 0.944  & 0.839  & 0.714  & 0.964  & 11.376  \\
          & Zheng Model  & 0.954  & 0.883  & 0.799  & 0.966  & 14.185  \\
          & Shahrian Model  & 0.912  & 0.906  & 0.899  & 0.913  & 145.860  \\
          & Improving Model & 0.954  & 0.883  & 0.799  & 0.966  & 381.923  \\
          & Proposed & 0.957  & 0.881  & 0.791  & 0.970  & 15.740  \\
    \midrule
    \multirow{5}[2]{*}{CHASE\_DB1} & Anat Model  & 0.944  & 0.820  & 0.675  & 0.964  & 29.047  \\
          & Zheng Model  & 0.944  & 0.820  & 0.675  & 0.964  & 40.587  \\
          & Shahrian Model  & 0.918  & 0.858  & 0.790  & 0.927  & 340.899  \\
          & Improving Model & 0.944  & 0.820  & 0.675  & 0.965  & 797.960  \\
          & Proposed & 0.951  & 0.815  & 0.657  & 0.973  & 50.710  \\
    \bottomrule
    \end{tabular}%
  \label{tab:comparedModel}%
\end{table}%

\subsection{Sensitivity analysis of threshold values of region features}
\begin{figure}
  \centering
  \subfigure[]{
	\includegraphics[width=1.625in,height=1.2in]{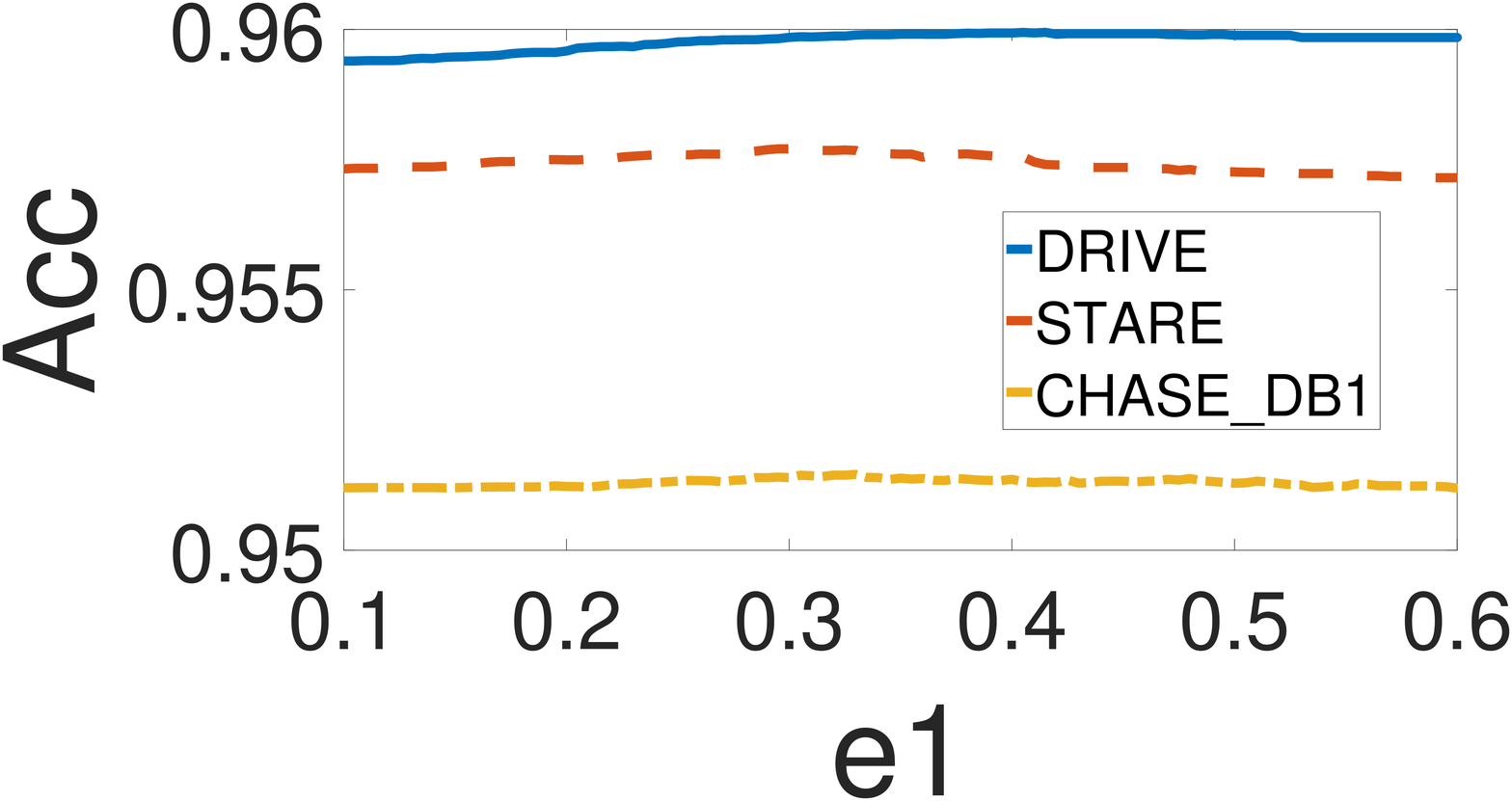}
  }
  \subfigure[]{
	\includegraphics[width=1.625in,height=1.2in]{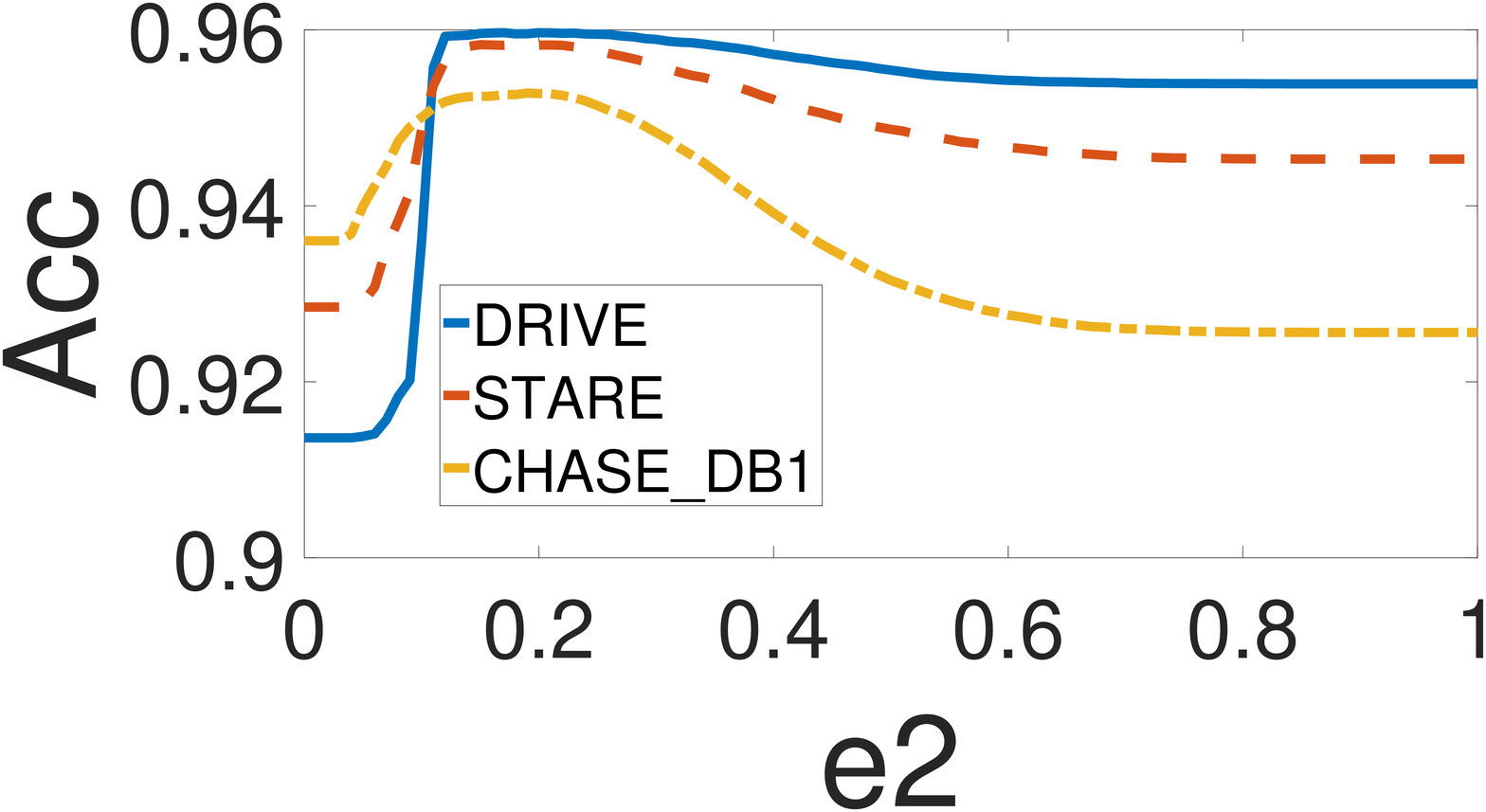}
  }
  \subfigure[]{
	\includegraphics[width=1.625in,height=1.2in]{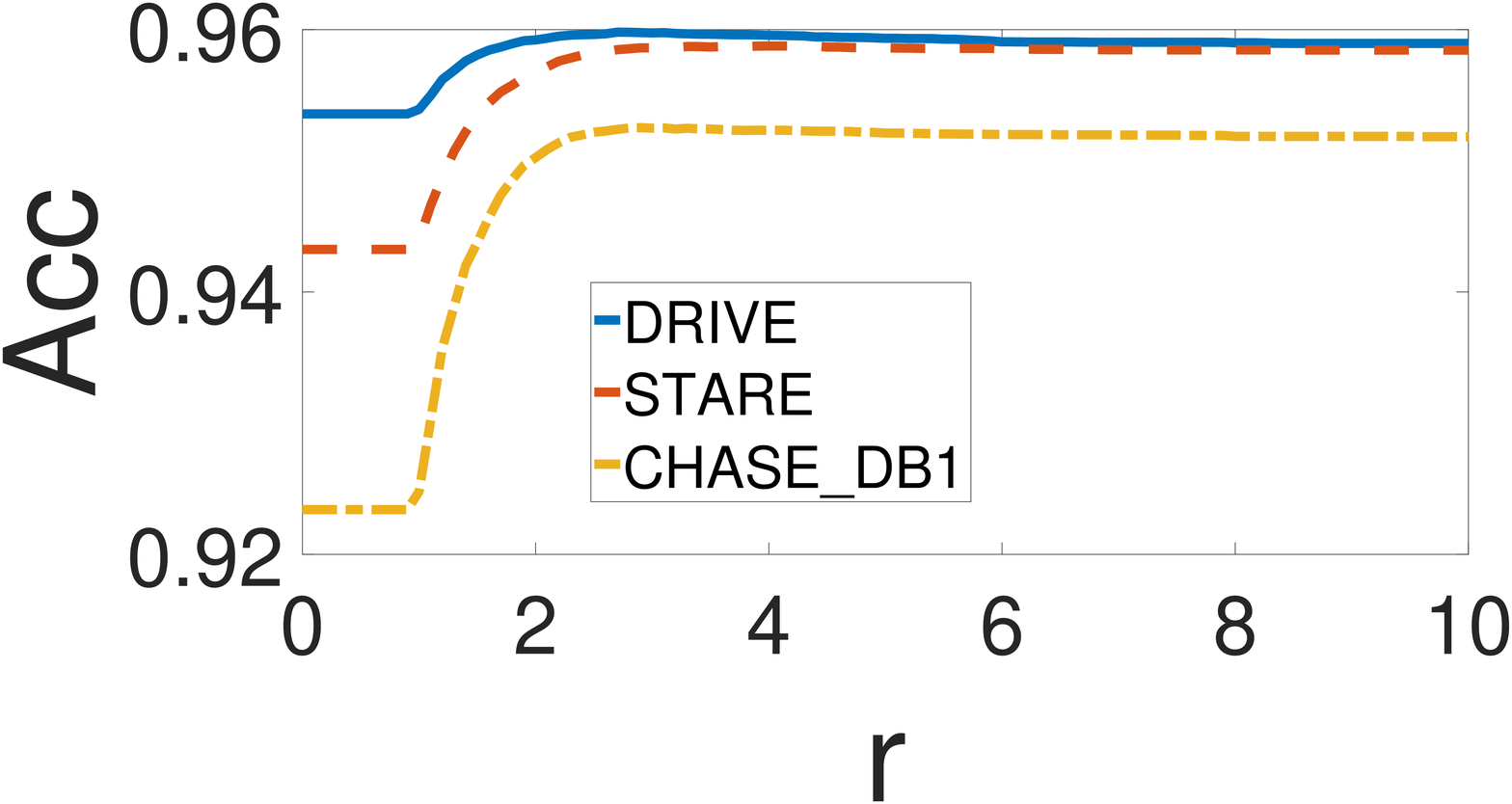}
  }
  \subfigure[]{
	\includegraphics[width=1.625in,height=1.2in]{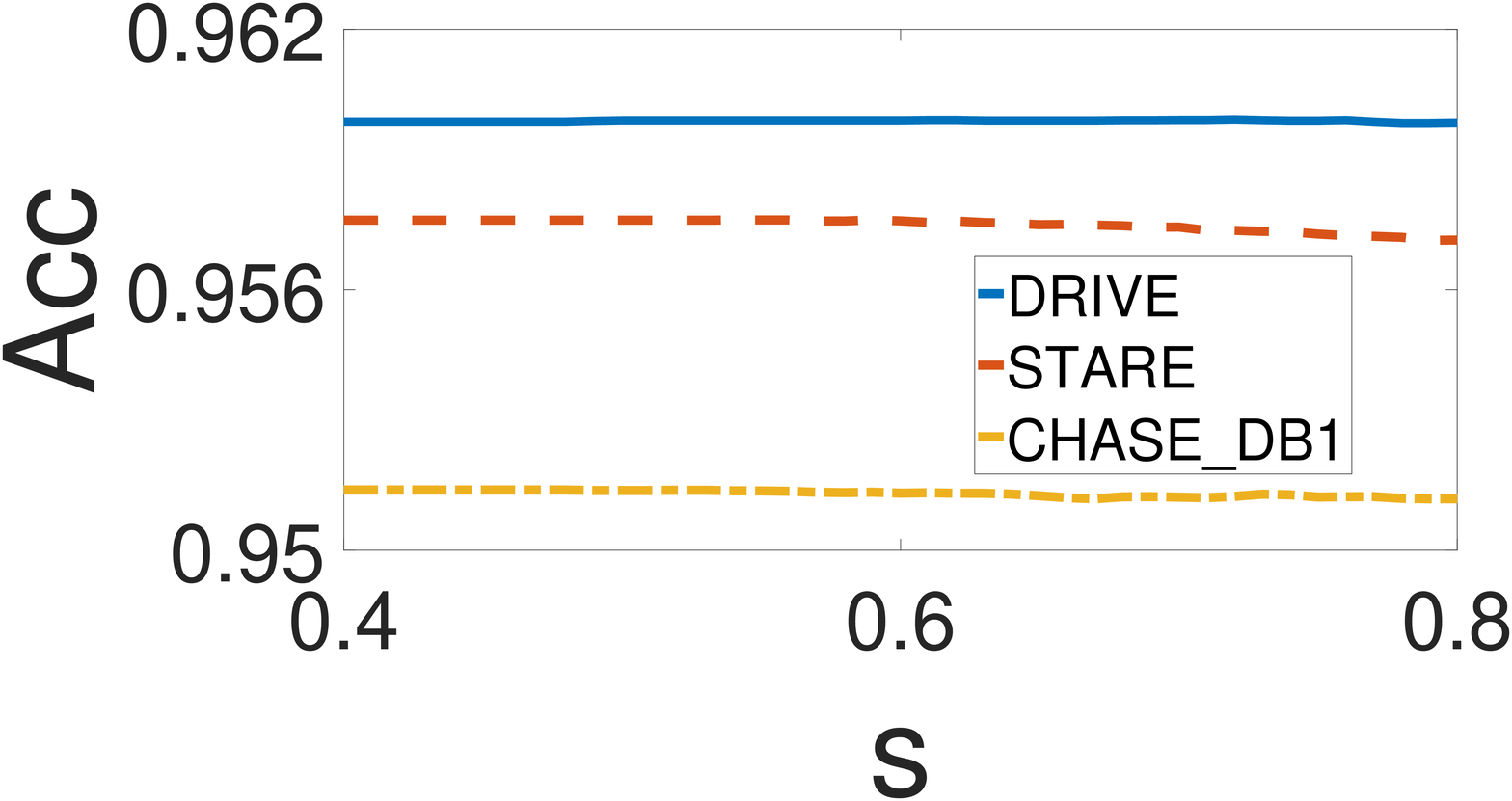}
  }

  \caption{Sensitivity analysis of threshold values of region features used in the work. (a) Variations in mean segmentation accuracy by varying $e_1$ when $r=2.2$, $s=0.53$, $e_2=0.25$. (b) Variations in mean segmentation accuracy by varying $e_2$ when $r=2.2$, $e_1=0.35$, $s=0.53$. (c) Variations in mean segmentation accuracy by varying $r$ when $e_1=0.35$, $s=0.53$, $e_2=0.25$. (d) Variations in mean segmentation accuracy by varying $s$ when $r=2.2$, $e_1=0.35$, $e_2=0.25$.}
  \label{fig:parameter}
\end{figure}

The default threshold values of region features: $e_1=0.35$, $r=2.2$, $s=0.53$, $e_2=0.25$ are used in our work. In order to demonstrate the insensitivity of the proposed model to these threshold values, the variations in $Acc$ by varying $e_1$, $e_2$, $r$ and $s$ are given in Fig.\ref{fig:parameter}.(a), (b), (c) and (d). From Fig.\ref{fig:parameter}, it can be observed that the proposed model can maintain high segmentation accuracy on the DRIVE, STARE and CHASE\_DB1 datasets as $e_1$ varies in $[0.2,0.4]$ or $e_2$ varies in $[0.15,0.3]$; For $r$ and $s$, the proposed model can maintain high segmentation accuracy as $r$ varies in $[2,6]$ or $s$ varies in $[0.4,0.6]$. From the above observation, it can be seen that the proposed model is not sensitive to these threshold values of region features when they change in a relatively large range.

\section{Conclusion}
Image matting refers to the problem of accurately extracting a foreground object from an input image, which is very useful in many important applications. However, to the best of our knowledge, image matting has never been employed before to extract blood vessels from fundus image. The major reason is that for retinal blood vessel segmentation, generating a user specified trimap is a tedious and time-consuming task. In addition, a normal image matting model needs to be designed carefully to improve the performance of blood vessel segmentation. In order to overcome these problems, region features of blood vessels are applied to generate the trimap automatically. Then a hierarchical image matting model is proposed to extract the pixels of blood vessel from the unknown regions. More specifically, a hierarchical strategy utilizing the continuity and extendibility of retinal blood vessels is integrated into the image matting model for blood vessel segmentation.

The proposed model is very efficient and effective in blood vessel segmentation, which achieves a segmentation accuracy of $96.0\%$, $95.7\%$ and $95.1\%$ on three public available datasets with an average time of $10.72s$, $15.74s$ and $50.71s$, respectively. The experimental results show that it is a very competitive model compared with many other segmentation approaches, and has a low computational time.

\ifCLASSOPTIONcaptionsoff
  \newpage
\fi

\bibliographystyle{IEEEtran}
\bibliography{VesselSegmentation}

\end{document}